\documentclass[10pt,journal,compsoc]{IEEEtran}

\usepackage{array}
\usepackage{hyperref}
\usepackage{url}
\usepackage[cmex10]{amsmath}
\usepackage{amssymb,amsthm}
\usepackage{epsfig,graphics}
\usepackage{multicol,subfigure}
\usepackage{wrapfig, rotating}
\usepackage{algorithmic,algorithm}
\usepackage{amsthm}

\usepackage[mathscr]{eucal}              
\usepackage{amsbsy}                      
\usepackage{bm}                          
\usepackage{paralist}                    
\usepackage{xspace}                      
\usepackage{caption}

\newcommand{\Sec}[1]{\hyperref[sec:#1]{\S\ref*{sec:#1}}} 
\newcommand{\Eqn}[1]{\hyperref[eq:#1]{(\ref*{eq:#1})}} 
\newcommand{\Fig}[1]{\hyperref[fig:#1]{Figure~\ref*{fig:#1}}} 
\newcommand{\Tab}[1]{\hyperref[tab:#1]{Table~\ref*{tab:#1}}} 
\newcommand{\Thm}[1]{\hyperref[thm:#1]{Theorem~\ref*{thm:#1}}} 
\newcommand{\Lem}[1]{\hyperref[lem:#1]{Lemma~\ref*{lem:#1}}} 
\newcommand{\Prop}[1]{\hyperref[prop:#1]{Property~\ref*{prop:#1}}} 
\newcommand{\Cor}[1]{\hyperref[cor:#1]{Corollary~\ref*{cor:#1}}} 
\newcommand{\Def}[1]{\hyperref[def:#1]{Definition~\ref*{def:#1}}} 
\newcommand{\Alg}[1]{\hyperref[alg:#1]{Algorithm~\ref*{alg:#1}}} 
\newcommand{\Ex}[1]{\hyperref[ex:#1]{Example~\ref*{ex:#1}}} 

\newcommand{\Tra}{^{\top}} 

\newcommand{\V}[1]{{\bm{\mathbf{\MakeLowercase{#1}}}}} 

\newcommand{\M}[1]{{\bm{\mathbf{\MakeUppercase{#1}}}}} 

\newcommand{\T}[1]{\boldsymbol{\mathscr{\MakeUppercase{#1}}}} 

\newcommand{\tube}[1]{\boldsymbol{\dot{\mathfrak{\MakeLowercase{#1}}}}} 
\newcommand{\tcol}[1]{\boldsymbol{\mathring{\mathfrak{\MakeLowercase{#1}}}}} 
\newcommand{\tvec}[1]{\boldsymbol{\vec{\mathfrak{\MakeLowercase{#1}}}}} 

\newcommand{\TA}{\T{A}}
\newcommand{\TB}{\T{B}}
\newcommand{\TS}{\T{S}}
\newcommand{\TC}{\T{C}}
\newcommand{\TU}{\T{U}}
\newcommand{\TV}{\T{V}}

\newcommand{\fft}{ \mbox{\tt fft} }
\newcommand{\ifft}{ \mbox{\tt ifft} }

\newtheorem{theorem}{Theorem}[section]
\numberwithin{theorem}{section}
\newtheorem{lemma}{Lemma}[section]
\numberwithin{lemma}{section}
\newtheorem{defn}{\indent \bf Definition}[section]
\numberwithin{defn}{section}

\numberwithin{conjecture}{section}
\newtheorem{remark}{Remark}[section]
\numberwithin{remark}{section}
\newtheorem{proposition}{Proposition}[section]
\numberwithin{proposition}{section}
\newtheorem{fact}{Fact}[section]
\numberwithin{fact}{section}

\makeatletter
\newsavebox\myboxA
\newsavebox\myboxB
\newlength\mylenA

\newcommand*\xbar[2][0.75]{%
    \sbox{\myboxA}{$\m@th#2$}%
    \setbox\myboxB\null
    \ht\myboxB=\ht\myboxA%
    \dp\myboxB=\dp\myboxA%
    \wd\myboxB=#1\wd\myboxA
    \sbox\myboxB{$\m@th\overline{\copy\myboxB}$}
    \setlength\mylenA{\the\wd\myboxA}
    \addtolength\mylenA{-\the\wd\myboxB}%
    \ifdim\wd\myboxB<\wd\myboxA%
       \rlap{\hskip 0.5\mylenA\usebox\myboxB}{\usebox\myboxA}%
    \else
        \hskip -0.5\mylenA\rlap{\usebox\myboxA}{\hskip 0.5\mylenA\usebox\myboxB}%
    \fi}
\makeatother

\begin{document}

\title{Exact tensor completion using t-SVD}
\author{Zemin~Zhang,~\IEEEmembership{Student Member,~IEEE,}
        Shuchin~Aeron,~\IEEEmembership{Member,~IEEE,} 
\IEEEcompsocitemizethanks{\IEEEcompsocthanksitem Zemin Zhang and Shuchin Aeron are with the Department
of Electrical and Computer Engineering, Tufts University, Medford, MA, 02155.\protect\\
E-mail: zemin.zhang@tufts.edu, shuchin@ece.tufts.edu\protect\\
\textbf{This work is supported by the NSF grant CCF:1319653} }}

\markboth{IEEE TRANSACTIONS ON PATTERN ANALYSIS AND MACHINE INTELLIGENCE, VOL.0, NO.0, 2015}%
{Shell \MakeLowercase{\textit{et al.}}: Tensor Completion using t-SVD: Theory and methods}

\IEEEtitleabstractindextext{%
\begin{abstract}
In this paper we focus on the problem of completion of multidimensional arrays (also referred to as tensors) from limited sampling. Our approach is based on a recently proposed tensor-Singular Value Decomposition (t-SVD) \cite{tensor_as_operator_Kilmer}. Using this factorization one can derive notion of tensor rank, referred to as the tensor tubal rank, which has optimality properties similar to that of matrix rank derived from SVD. As shown in \cite{Zhang:2014:NMM:2679600.2680311} some multidimensional data, such as panning video sequences exhibit low tensor tubal rank and we look at the problem of completing such data under random sampling of the data cube. We show that by solving a convex optimization problem, which minimizes the tensor nuclear norm obtained as the convex relaxation of tensor tubal rank, one can guarantee recovery with overwhelming probability as long as samples in proportion to the degrees of freedom in t-SVD are observed. In this sense our results are order-wise optimal. The conditions under which this result holds are very similar to the incoherency conditions for the matrix completion, albeit we define incoherency under the algebraic set-up of t-SVD. We show the performance of the algorithm on some real data sets and compare it with other existing approaches based on tensor flattening and Tucker decomposition.
\end{abstract}

}

\maketitle
\IEEEdisplaynontitleabstractindextext
\IEEEpeerreviewmaketitle


\section{Introduction}
\label{sec:1}
\IEEEPARstart{D} ata in the form of multidimensional array of numbers, also referred to as \emph{tensors}\footnote{\textbf{This terminology comes from representation of multilinear functionals on the outer product of finite dimensional vector spaces as an indexed array of numbers, \cite{Hackbusch_book}}}, arises naturally in a number of scenarios. Recovery of tensors under limited number of measurements is an important problem, which arises in a variety of applications, such as recommendation systems \cite{boumalabsil2011rtrmc}, dimensionality reduction \cite{DBLP:journals/combinatorica/LinialLR95}, multi-class learning \cite{DBLP:journals/sac/ObozinskiTJ10},data mining \cite{4781131,DBLP:conf/sdm/SunPLCLQ09} and computer vision \cite{6138863,Zhang:2014:NMM:2679600.2680311}. 



The strategies for sampling and recovery of tensors rest heavily on the framework used to reveal an algebraic structure in the data, namely a \emph{low-rank} factorization. For example for the widely studied case of matrix (an order-$2$ tensor) completion, a random sampling strategy, which  is \emph{incoherent} with respect to the left and right singular vectors in the SVD, have shown to be nearly optimal in terms of sampling complexity. Moreover it is shown that a computationally feasible method based on minimizing the nuclear norm (as a convex surrogate for low rank) is sufficient for recovery \cite{exact_matrix_completion_candes_benj,simpler_approach_benj}. For tensors of order higher than $2$, the fundamental algebraic factorization namely CANDECOMP/PARAFAC(CP) \cite{Kolda09tensordecompositions}, which decomposes a tensor as a sum of rank-1 factors can be used. However there are known computational and ill-posedness issues with CP \cite{LH}. Other kinds of decompositions such as Tucker, Hierarchical-Tucker (H-Tucker) and Tensor Train (TT) \cite{Grasedyck:2010} are also shown to reveal the algebraic structure in the data with the notion of rank extended to the notion of a multi-rank, expressed as a vector of ranks of the factors in the contracted representation using matrix product states. Approaches based on Tucker decomposition work by \emph{unfolding} the tensor in various ways, subsequently employing the theory and methods for matrix completion, see for example \cite{BHuangCmu,GandyRY2011}. 

In this paper we consider sampling and recovery for 3-rd order tensors using the tensor-SVD (t-SVD) proposed in \cite{tensor_as_operator_Kilmer} as a rank-revealing factorization and exploited in \cite{Zhang:2014:NMM:2679600.2680311} for problem of video recovery from missing pixels. In this paper we derive theoretical performance bounds for recovery under the setting of \cite{Zhang:2014:NMM:2679600.2680311}. t-SVD is essentially based on a group theoretic approach where the multidimensional structure is unraveled by constructing group-rings along the tensor fibers \cite{Navasca_2010}\footnote{\textbf{In this paper, in interest of clarity of exposition, we restrict ourselves to group rings constructed out of cyclic groups, resulting in an algebra of circulants \cite{Gleich}. Nevertheless, the results presented here hold true for the general group-ring construction and the algorithms are also directly applicable.}}. The advantage of such an approach over the existing approaches is that the resulting algebra and analysis is very close to that of matrix algebra and analysis. In particular, our work has been greatly inspired by \cite{simpler_approach_benj} and it turns out the main tool namely the Non-commutative Bernstein Inequality (NBI) is also helpful in deriving our results. We prove that we can perfectly recover a tensor of size $n_1 \times n_2 \times n_3$ with rank $r$ under t-SVD, also referred to as the tubal-rank (see Section \ref{sec:2}), by solving a convex optimization problem, given ${\cal O}(r n_1 n_3 \log((n_1+n_2)n_3) ) $ samples. In order to highlight our contributions we now go over related work on tensor completion using different tensor factorizations and contrast our findings with existing literature.

\subsection{Related Work}
\begin{table*}
	\caption{A summary of existing tensor completion methods}
	\begin{center}
		\begin{tabular}{|p{3cm}|p{4cm}|p{4.5cm}|p{4.5cm}|}
			\hline
			\textbf{Format} & \textbf{Sampling Method} & \textbf{Samples needed for exact recovery (3rd-order tensor of size $n \times n \times n$)}  & \textbf{Incoherent condition} \\ \hline
			CP\cite{DBLP:conf/icml/MuHWG14}  & Gaussian measurements & $O(rn^{2})$ for CP rank $r$ & N/A  \\ \hline
			CP\cite{NIPS2014_5475}  & Random sampling     & $O(n^{3/2}r^5 \log^4(n))$ for CP rank $r$ on symmetric tensors & Incoherent condition of symmetric tensors with orthogonal decomposition \\ \hline
			Tucker\cite{DBLP:conf/icml/MuHWG14} & Gaussian measurements& $O(rn^2)$ for Tucker rank $(r,r,r)$ & N/A \\ \hline
			Tucker\cite{BHuangCmu} & Random downsample & $O(rn^2 \log^2(n))$ for Tucker rank $(r,r,r)$ & Matrix incoherent condition on all mode-$n$ unfoldings \\ \hline
			CP\cite{conf/nips/KrishnamurthyS13} & Adaptive Sampling & $O(nr\log(r))$ for CP rank r & Standard incoherent condition with orthogonal decomposition \\ \hline
			t-SVD(this paper)  & Random sampling & $O(rn^2 \log(n^2))$ for Tensor tubal-rank $r$ & Tensor incoherent condition \\
			\hline
		\end{tabular}
	\end{center}
    \label{table}
\end{table*}

Apart from t-SVD, there are mainly two types of low-rank tensor completion methods considered in the literature, methods that are based on the CP format, and those that are based on the Tucker decomposition. The sampling methods include random downsampling, Gaussian measurements and adaptive sampling. We summarize these results in \textbf{Table}~\ref{table}. Below we will provide details for each of these methods. 


\subsubsection{Tensor Completion Based on CP decomposition}
 The CP decomposition of a tensor $\T{X} \in \mathbb{R}^{I_1 \times I_2 \times ... \times I_N}$ of order $N$ is given by,
\begin{equation}
\label{CP format}
\T{X} = \sum\limits_{\ell=1}^{L} \V{x}_\ell^{(1)} \circ \V{x}_\ell^{(2)} \circ \dots \circ \V{x}_\ell^{(N)},\hspace{2mm} \V{x}_\ell^{(n)} \in \mathbb{R}^{I_n} \,\, ,
\end{equation}
where $\circ$ denotes the vector outer product \cite{Kolda09tensordecompositions}. The smallest $L$ in such that Equation~(\ref{CP format}) holds is called the CP rank of $\T{X}$. 

Suppose we sample $\T{X}$ at the set of indices in a set $\Omega$. Let $P_\Omega$ the orthogonal projection onto $\Omega$. Then \cite{Karlsson.Kressner.Uschmajew:2014} tries to complete the tensor by solving the following optimization problem,
\begin{equation}
\min \|P_\Omega(\T{A} - \sum\limits_{\ell=1}^{L} \V{x}_\ell^{(1)} \circ \V{x}_\ell^{(2)} \circ ... \circ \V{x}_\ell^{(N)} ) \|^2_2 + \lambda \sum\limits_{\ell=1}^{L} \sum\limits_{n=1}^{N}\|\V{x}_\ell^{(n)}\|_2^2\,\, ,
\end{equation}
where $\lambda \ge 0$ is the regularization parameter. However, this approach has several drawbacks. The optimization problem is non-convex and hence only local minima can be guaranateed. Further, for practical problems it is often computationally difficult to determine the CP rank or the best low rank CP approximation of a tensor data beforehand.

Recently in \cite{NIPS2014_5475} the authors show that they can recover an $n \times n \times n$ \emph{symmetric} tensor with CP rank $r$ exactly from $O(n^{3/2}r^5\log^4 n)$ randomly sampled entries under standard incoherency conditions on the factors. A tensor $\T{X} \in \mathbb{R}^{n \times n \times n}$ is called symmetric in the CP format if its CP decomposition has the format $\T{X} = \sum\limits_{\ell=1}^{r} \sigma_\ell (\V{u}_\ell \circ \V{u}_\ell \circ \V{u}_\ell)$, where $\V{u}_\ell \in \mathbb{R}^n$ with $\|\V{u}_\ell\| = 1$. They proved that their methods yield global convergence. Although the authors stressed that their results generalize easily to handle non-symmetric tensors and high-order tensors, there is no explicit result supporting this claim and it is unclear how to extend the result to general tensors.

\subsubsection{Tensor Completion Based on Tucker Decomposition}
Liu et al. \cite{10.1109/TPAMI.2012.39} proposed tensor completion based on minimizing tensor $n$-rank in Tucker Decomposition format, which is the matrix rank of mode-$n$ matricization of a tensor.
Specifically, using the matrix nuclear norm instead of matrix rank, they are trying to solve the convex problem as follows:

\begin{equation}
\begin{aligned}
\label{Tensor n rank}
\min_{\T{X}} &\sum\limits_{i=1}^{n}\alpha_i \|\T{X}_{(i)}\|_*\\
\mbox{subject to  } &P_\Omega(\T{X}) = P_\Omega(\T{T}) 
\end{aligned}
\end{equation}
where $\T{X}_{(i)}$ is the mode-$n$ matricization of $\T{X}$ and $\alpha_i$ are prespecified constants satisfying $\alpha_i \ge 0$ and $\sum_{i=1}^{n} \alpha_i = 1$. It also has several drawbacks. There is no specific way on how to choose weights $\alpha_i$ and normally one just chooses one best matricization according to the results, which turns to a matrix completion problem. It is not applicable to higher-order tensors since no one knows how to assign each $\alpha_i$. Also there is no theoretical guarantees of how many number of samples are required for exact recovery.

There are some modified versions of this problem. \cite{DBLP:conf/nips/Romera-ParedesP13} changed the above complexity term to the convex envelop of cardinality function of singular value vectors, which is proved to be a tighter convex relaxation of the tensor $n$-rank compared to the above approach. The approach uses the Alternating Direction Method of Multipliers (ADMM) for global convergence guarantees. 

Huang et al. \cite{BHuangCmu} solved the following convex program, which can be regarded as the combination of the matrix completion and the matrix Robust Principal Component Aanlysis (RPCA), when extended to the case of tensors,
\begin{equation}
\begin{aligned}
\label{tensor n rank 2}
\min_{\T{X},\T{E}} &\sum\limits_{i=1}^{K} \alpha_i \|\T{X}_{(i)}\|_* + \|\T{E}\|_1 +\frac{\tau}{2}\|\T{X}\|^2_F + \frac{\tau}{2} \|\T{E}\|^2_F \\
\mbox{s.t.  } &P_\Omega(\T{X}+\T{E}) = \T{B}  
\end{aligned}
\end{equation}
 They outlined the Tensor Incoherent Conditions (TIC) of exact recovery of tensors using (\ref{tensor n rank 2}) with $\alpha_i = \sqrt{n_i^{(1)}}$, where $n_i^{(1)} = \max \{n_i,\prod_{j \neq i}n_j \}$. It is essentially a direct extension of matrix completion since for each mode-$n$ matricization of the tensor they applied the matrix incoherent condition and get a bound for each mode. The number of samples needed for exact recovery also have the same form as matrix completion result. A drawback of this approach is that the tensor needs to satisfy the TIC in the unfoldings of every mode. 

In \cite{DBLP:conf/icml/MuHWG14} Mu et al. proposed a different method for tensor completion for a tensor with low Tucker rank under Gaussian measurements instead of random sampling. The core idea is to reshape a multi-dimensional tensor $\T{X} \in \mathbb{R}^{n_1 \times n_2 \times n_3 \times ... \times n_K}$ into a matrix $\T{X}_{(j)} \in \mathbb{R}^{n_1 ... n_j \times n_{j+1}...n_{K}}$ and apply matrix completion on it. It can be shown easily if $\T{X}$ is a low-rank tensor (in either CP or Tucker sense), $\T{X}_{(j)}$ will be a low-rank matrix. They showed if $\T{X}_0$ has CP rank $r$, then $m \ge Crn^{\lceil K/2 \rceil}$ will be sufficient to recover the original tensor. If $\T{X}_0$ has Tucker rank $(r,r,...,r)$, then $m \ge Cr^{\lfloor K/2 \rfloor } n^{\lceil K/2 \rceil }$ is sufficient. As we will show later on, if a third order tensor $\T{X}$ is of CP rank $r$, then its tensor tubal rank is also at most $r$. So in this sense it means for any third-order tensor with low CP rank, we can recover it from random samples using this t-SVD structure. Further our technique can be extended to higher-order tensors very easily. Given an $n_1 \times n_2 \times ... \times n_K$ tensor one can reshape it to form a third-order tensor of size $(n_1n_2..n_{k_1}) \times (n_{k_1+1}..n_{K-1}) \times n_K$ and then use our framework for 3-D tensor completion. If $n_k = n$ for all $K$ dimensions then it follows from our main results that $O(n^{\lceil K/2 \rceil} r \log (n^{\lceil K/2 \rceil} )$ measurements are sufficient enough for recovery.

\subsubsection{Tensor Completion via Adaptive Sampling}
Krishnamurthy and Singh \cite{conf/nips/KrishnamurthyS13} develop a tensor completion approach based on the adaptive sampling. The key idea is to predict the tensor singular subspace given the sampled sub-tensor, and recursively update the subspace if a newly sampled sub-tensor lies out of it. They proved that $O(nr^{3/2}\log r)$ adaptively chosen samples are sufficient for exact recovery when the tensor is of CP rank $r$. This approach extends the matrix completion method to the tensor case and yield a tighter bound with only column incoherency conditions.

\subsection{Organization of the paper}
This paper is organized as follows. Section~\ref{sec:2} introduces some notations and preliminaries of tensors, where we defined several algebraic structures of third-order tensors. In Section~\ref{sec:3} we derive the main result on tensor completion and give out the provable bound using t-SVD. We provide the full proof in Section~\ref{sec:4} and we report empirical and numerical results in Secion~\ref{sec:5} . Finally we outline implications of the work and future research in Section~\ref{sec:6}.

\section{Notations and preliminaries}
\label{sec:2}
Before we going to the main result, we need to go over the notations used in this paper. Some of the notations and definitions of tensors are discussed in \cite{KBN,tensor_as_operator_Kilmer,Zhang:2014:NMM:2679600.2680311} and some notations are novel in this section.

Tensors are represented in bold script font. For instance, a third-order tensor is represented as $\T{A}$, and its $(i,j,k)th$ entry is represented as $\T{A}_{ijk}$. Moreover, a tensor tube of size $1 \times 1 \times n_3$ is denoted as $\tube{a}$, and a tensor column of size $n_1 \times 1 \times n_3$ will be $\tcol{b}$. $\widehat{\T{A}}$ is a tensor which is obtained by taking the Fast Fourier Transform (FFT) along the third mode of $\T{A}$. For a compact notation we will use the following convention for Fourier transform along the 3rd dimension-  $\widehat{\T{A}} = \fft(\T{A},[\hspace{1mm}],3)$. In the same fashion, one can also compute $\T{A}$ from $\widehat{\T{A}}$ via $\ifft(\widehat{\TA},[\hspace{1mm}],3)$ using the inverse FFT.\\

The $i$th frontal slice of $\TA$ is defined as $\TA^{(i)}$. Frequently we will also use the following notation $\TA^{(i)} = \TA(:,:,i)$ to denote fixing of a particular index and collecting the particular slice/fiber of the tensor. Similarly, $\widehat{\TA}^{(i)}$ is the $i$th frontal slice of $\widehat{\TA}$.

\begin{defn} \textbf{(Tensor transpose)}
The conjugate transpose of a tensor $\TA \in \mathbb{R}^{n_1 \times n_2 \times n_3}$ is the $n_2 \times n_1 \times n_3$ tensor $\TA\Tra$ obtained by conjugate transposing each of the frontal slice and then reversing the order of transposed frontal slices $2$ through $n_3$.
\end{defn} 

\begin{figure}[htbp]
\centering \makebox[0in]{
    \begin{tabular}{c c}
      \includegraphics[scale=0.35]{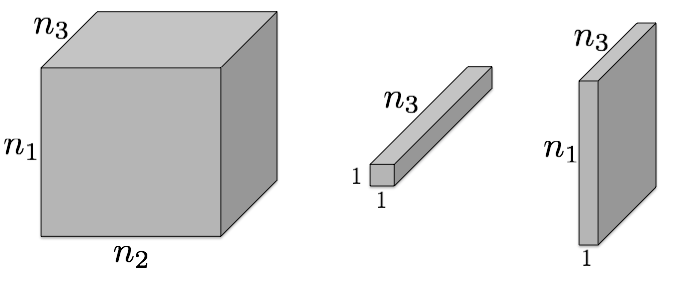}
 \end{tabular}}
  \caption{ General third-order tensor, tensor tube and tensor column.}
  \label{fig:tensor}
\end{figure}

One can define the t-product between two tensors by introducing a commutative operation, which is circular convolution $\star$ in our case, as follows.
 
\begin{defn}\textbf{(t-product)}
The t-product $\T{A}*\T{B}$ of $\TA \in \mathbb{R}^{n_1 \times n_2 \times n_3}$ and $\TB \in \mathbb{R}^{n_2 \times n_4 \times n_3}$ is an $n_1 \times n_4 \times n_3$ tensor whose $(i,j)$th tube $\tube{c}_{ij}$ is given by
\begin{equation}
\tube{c}_{ij} = \TC(i,j,:) = \sum\limits_{k=1}^{n2}\TA(i,k,:) \star \TB(k,j,:)
\end{equation}
where $\star$ denotes the circular convolution between two tubes of same size. 
\end{defn}

A 3-D tensor of size $n_1 \times n_2 \times n_3$ can be viewed as an $n_1 \times n_2$ matrix of tubes which lie in the third-dimension. So the t-product of two tensors can be regarded as a matrix-matrix multiplication, except for the operation between numbers is replaced by circular convolution between the tubes.

\begin{defn}\textbf{(Identity tensor)}
The identity tensor $\T{I} \in \mathbb{R}^{n_1 \times n_1 \times n_3}$ is defined to be a tensor whose first frontal slice $\T{I}^{(1)}$ is the $n_1 \times n_1$ identity matrix and all other frontal slices $\T{I}^{(i)},i=2,...,n_3$ are zero.\\
\end{defn}
\begin{defn}\textbf{(Orthogonal tensor)}\cite{tensor_as_operator_Kilmer}
A tensor $\T{Q} \in \mathbb{R}^{n \times n \times n_3}$ is orthogonal if it satisfies
\begin{equation}
\T{Q}\Tra * \T{Q} = \T{Q} * \T{Q}\Tra = \T{I}
\end{equation}
\end{defn}

\begin{defn}\textbf{(Block diagonal form of third-order tensor)}
Let $\xbar{\T{A}}$ denote the block-diagonal matrix of the tensor $\TA\Tra$ in the Fourier domain, i.e.,
\begin{equation}
\begin{aligned}
\xbar{\T{A}} \triangleq &\text{ blockdiag}( \widehat{\T{A}} )\\
 \triangleq &\left[\begin{array}{cccc}\widehat{{\T{A}}}^{(1)}& & & \\
 & \widehat{{\T{A}}}^{(2)} & & \\
 & &\ddots & \\
  & & & \widehat{{\T{A}}}^{(n_3)}\end{array} \right] 
  \in \mathbb{C}^{n_1n_3 \times n_2n_3}
\end{aligned}
\end{equation} 

It is easy to verify that the block diagonal matrix of $\TA\Tra$ is equal to the transpose of the block diagonal matrix of $\TA$:
\begin{equation}
\xbar{\TA\Tra} = \xbar{\TA}\Tra 
\end{equation}
\end{defn}


\begin{remark}
The following fact will be used through out the paper. For any tensor $\TA \in \mathbb{R}^{n_1 \times n_2 \times n_3}$ and $\TB \in \mathbb{R}^{n_2 \times n_4 \times n_3}$, we have
\begin{equation}
\nonumber
\TA*\TB=\TC \Longleftrightarrow \xbar{\TA}\xbar{\TB} = \xbar{\TC}
\end{equation}

\end{remark}

The t-product allows us to define a tensor Singular Value Decomposition (t-SVD). We need one more definition to state the decomposition. 
 
\begin{defn} \textbf{(f-diagonal tensor)}
A tensor $\TA$ is called f-diagonal if each frontal slice $\TA^{(i)}$ is a diagonal matrix.
\end{defn}


\begin{defn}\textbf{(Tensor Singular Value Decomposition: t-SVD)}
For $\T{M}\in\mathbb{R}^{n_1 \times n_2 \times n_3}$, the t-SVD of $\T{M}$ is  given by
\begin{equation}
\T{M} = \T{U} *\T {S} *\T{V}\Tra
\end{equation}
\noindent where $\T{U}$ and $\TV$ are orthogonal tensors of size $n_1 \times n_1 \times n_3 $ and  $n_2 \times n_2 \times n_3 $ respectively. $\T{S}$ is a rectangular $f$-diagonal tensor of size $n_1 \times n_2 \times n_3 $, and the entries in $\T{S}$ are called the singular values of $\T{M}$. $*$ denotes the t-product here.
\end{defn}

One can obtain this decomposition by computing matrix SVDs in the Fourier domain, see Algorithm 1, where the the algorithm is given for a general order $N$ tensor, obtained by recursively applying the Fourier transform over successive dimensions \cite{dmartin:an}. Figure \ref{fig:tSVD} illustrates the decomposition for the 3-D case.
\begin{algorithm}
\label{alg:tSVD - for $N$-D tensor}
  \caption{t-SVD}
  \begin{algorithmic}
  \STATE \textbf{Input: } $\T{M} \in \mathbb{R}^{n_1 \times n_2  ... \times n_N}$
  \STATE $L = n_3n_4...n_N$
  \STATE $\T{D} = \T{M}$
   \FOR{$i = 3 \hspace{2mm} \rm{to} \hspace{2mm} N$}
  	\STATE ${\T{D}} \leftarrow \rm{fft}(\T{D},[\hspace{1mm}],i)$;
  \ENDFOR
  
  \FOR{$i = 1 \hspace{2mm} \rm{to} \hspace{2mm} L$}
  	\STATE $ [\M{U}, \M{S}, \M{V}] = {\tt svd}(\T{D}(:,:,i))$
  	\STATE $ {\hat{\T{U}}}(:,:,i) = \M{U}; \hspace{1mm} {\hat{\T{S}}}(:,:,i) = \M{S}; \hspace{1mm} {\hat{\T{V}}}(:,:,i) = \M{V}; $
  \ENDFOR
  
  \FOR{$i = 3 \hspace{2mm} \rm{to} \hspace{2mm} N$}
  	\STATE $\T{U} \leftarrow \rm{ifft}(\hat{\T{U}},[\hspace{1mm}],i); \hspace{1mm} \T{S} \leftarrow \rm{ifft}(\hat{\T{S}},[\hspace{1mm}],i); \hspace{1mm} \T{V} \leftarrow \rm{ifft}(\hat{\T{V}},[\hspace{1mm}],i)$;
  \ENDFOR
  \end{algorithmic}
\end{algorithm}

\begin{figure}[htbp]
\centering \makebox[0in]{
    \begin{tabular}{c c}
      \includegraphics[scale=0.45]{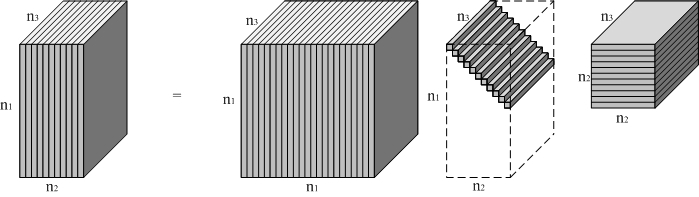}
 \end{tabular}}
  \caption{ The t-SVD of an $n_1 \times n_2 \times n_3$ tensor.}
  \label{fig:tSVD}
\end{figure}


Based on t-SVD one can define several notions of rank. 

\begin{defn} \textbf{(Tensor multi rank and tubal rank)}
The tensor multi rank of a tensor $\T{A} \in \mathbb{R}^{n_1 \times n_2 \times n_3}$ is a vector $\vec{r} \in \mathbb{R}^{n_3 \times 1}$ with the $i$th element equals to the rank of $i$th frontal slice of $\widehat{\T{A}}$ in Fourier domain. The tensor \textbf{tubal rank} $r$ of $\T{A}$ is defined to be the number of non-zero singular tubes of $\T{S}$, where $\T{S}$ comes from the t-SVD of $\T{A}: \T{A} = \TU *\TS* \TV\Tra$. An alternative definition of tubal rank is that it is the largest rank of all the frontal slices of $\hat{\TA}$, which means $r = \max \{ \vec{r}\}$. If we say a third order tensor $\T{A}$ is of \textbf{full rank}, it means $r = \text{min}\{n_1,n_2\}$.
\end{defn}

\begin{remark} It is usually sufficient to compute the reduced version of t-SVD using the tensor tubal-rank. It's faster and more economical for storage. In details, suppose $\T{M}\in \mathbb{R}^{n_1 \times n_2 \times n_3}$ has tensor tubal-rank $r$, then the reduced t-SVD of $\T{M}$ is given by
\begin{equation}
\T{M} = \T{U} *\T {S} *\T{V}\Tra
\end{equation}
\noindent where $\T{U} \in \mathbb{R}^{n_1 \times r \times n_3}$ and $\TV \in \mathbb{R}^{n_2 \times r \times n_3}$ satisfying $\TU\Tra * \TU = \T{I},\TV\Tra * \TV = \T{I}$. $\T{S}$ is a $f$-diagonal tensor of size $r \times r \times n_3 $. This reduced version of t-SVD will be used throughout the paper unless otherwise noted.
\end{remark}

\begin{remark}\textbf{(Relation to CP decomposition)}
Suppose a tensor $\TA \in \mathbb{R}^{n_1 \times n_2 \times n_3}$ has CP rank $r$ and its CP decomposition is given by 
\begin{equation}
\nonumber
\T{A} = \sum_{i=1}^{r} a_i^{(1)} \circ a_i^{(2)} \circ a_i^{(3)}
\end{equation}
where $a_i^{(k)} \in \mathbb{R}^{n_k},k=1,2,3$. Then we can see that the tensor $\widehat{\TA}$ which is obtained by taking the FFT along the third dimension of $\TA$, has the CP decomposition as follows,
\begin{equation}
\nonumber
\widehat{\T{A}} = \sum_{i=1}^{r} a_i^{(1)} \circ a_i^{(2)} \circ \widehat{a_i^{(3)}}
\end{equation}
where $\widehat{a_i^{(3)}} = \fft(a_i^{(3)}),i=1,2,...,r$. So $\widehat{\TA}$ also has CP rank $r$, and each frontal slice of $\widehat{\TA}$ is the sum of $r$ rank-$1$ matrices, so the rank of each frontal slice is at most $r$. This means if a tensor is of CP rank $r$, then its tensor tubal rank is at most $r$.
\end{remark}


\begin{defn}\textbf{(Inverse of tensor)}
The inverse of a tensor $\T{A} \in \mathbb{R}^{n \times n \times n_3}$ is written as $\T{A}^{-1}$ satisfying
\begin{equation}
\T{A}^{-1} * \T{A} = \T{A} * \T{A}^{-1} = \T{I}
\end{equation}
where $\T{I}$ is the \textbf{identity tensor} of size $n \times n \times n_3$.
\end{defn}


\begin{defn}
\textbf{(Tensor operator)}
Tensor operators are denoted by Calligraphic letters. Suppose $\mathfrak{L}: \mathbb{R}^{n_1 \times n_2 \times n_3} \rightarrow \mathbb{R}^{n_4 \times n_2 \times n_3} $ is a tensor operator mapping an $n_1 \times n_2 \times n_3$ tensor $\T{A}$ to an $n_4 \times n_2 \times n_3$ tensor $\T{B}$ via the t-product as follows:
\begin{equation}
\label{eq:operator}
\T{A} = \mathfrak{L}(\T{B}) = \T{L}*\T{B}
\end{equation}
where $\T{L}$ is an $n_4 \times n_1 \times n_3$ tensor. Then it is equivalent to the following equation, which lies in the Fourier domain of (\ref{eq:operator}):
\begin{equation}
\label{eq:operator_fourier}
\xbar{\TA} = \xbar{\T{L}} \xbar{\T{B}}
\end{equation}
where $\xbar{\T{A}} \in \mathbb{C}^{n_1n_3 \times n_2n_3}$, $\xbar{\T{L}} \in \mathbb{C}^{n_4n_3 \times n_1n_3}$ ,$\xbar{\T{B}} \in \mathbb{C}^{n_4n_3 \times n_2n_3}$ which are block diagonal matrices.
\end{defn}

\begin{remark}
 For a tensor operator via t-product defined in (\ref{eq:operator}), we are able to transform it into the equivalent form in Fourier domain (\ref{eq:operator_fourier}). On the other hand we can also transform an operator in Fourier domain back to the original domain as needed.
\end{remark}


\begin{defn}\textbf{(Inner product of tensors)}
If $\T{A}$ and $\T{B}$ are third-order tensors of same size $n_1 \times n_2 \times n_3$, then the inner product between $\T{A}$ and $\T{B}$ is defined as the following,
\end{defn}
\begin{equation}
\label{def:inner_product}
\langle \T{A} , \T{B} \rangle = \frac{1}{n_3}\text{trace}( \xbar{\TB}\Tra \xbar{\TA} ) \in \mathbb{R}
\end{equation}
where $1/n_3$ comes from the normalization constant of the FFT. The reason that this inner product produces a real-valued result comes from the conjugate symmetric property of the FFT.


\begin{defn}
\label{def:basis}
\textbf{(Standard tensor basis and the corresponding decomposition)}
We introduce 2 standard tensor basis here. The first one is called \textbf{column basis} $\tcol{e}_i$ of size $n \times 1 \times n_3$ with only one entry equaling 1 and the rest equaling zero. However, the nonzero entry 1 will only appear at the first frontal slice of $\tcol{e}_i$. Naturally its transpose $\tcol{e}_i\Tra$ is called \textbf{row basis}. The other standard tensor basis is called \textbf{tube basis} $\tube{e}_i$ of size $1 \times 1 \times n_3$ with one entry equaling to 1 and rest equaling to 0. Figure~\ref{fig:basis} illustrates these 2 basis.
\end{defn}

\begin{figure}[htbp]
\centering \makebox[0in]{
    \begin{tabular}{c c}
      \includegraphics[scale=0.45]{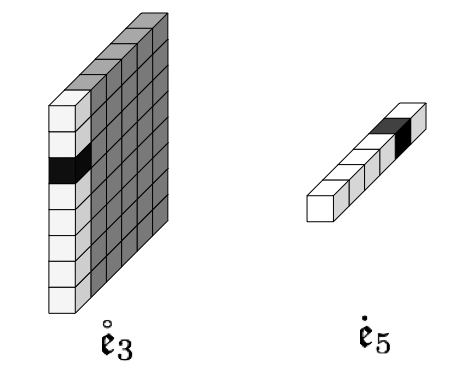}
 \end{tabular}}
  \caption{ The column basis $\tcol{e}_3$ and tube basis $\tube{e}_5$. The black cubes are 1, gray and white cubes are 0. The white cubes stand for the potential entries that could be 1.}
  \label{fig:basis}
\end{figure}

And one can obtain a unit tensor $\T{E}$ with the only non-zero entry $\T{E}_{ijk}$ equaling to 1 via the following:
\begin{equation}
\T{E} = \tcol{e}_i * \tube{e}_k * \tcol{e}_j\Tra
\end{equation}

Given any third order tensor $\T{X} \in \mathbb{R}^{n_1 \times n_2 \times n_3}$, we have the following decomposition
\begin{equation}
\begin{aligned}
\nonumber
\T{X} &= \sum\limits_{i=1}^{n_1} \sum\limits_{j=1}^{n_2} \sum\limits_{k=1}^{n_3} \langle \tcol{e}_i * \tube{e}_k * \tcol{e}_j\Tra , \T{X} \rangle \tcol{e}_i * \tube{e}_k * \tcol{e}_j\Tra\\
&= \sum\limits_{i=1}^{n_1} \sum\limits_{j=1}^{n_2} \sum\limits_{k=1}^{n_3} \T{X}_{ijk} \tcol{e}_i * \tube{e}_k * \tcol{e}_j\Tra
\end{aligned}
\end{equation}
The proof to such a decomposition is straightforward since $\langle \tcol{e}_i * \tube{e}_k * \tcol{e}_j\Tra , \T{X} \rangle$ will give out the exact value of $\T{X}_{ijk}$.

Variety of novel norms defined on tensors will be defined in the following. These norms will be used throughout the proof of our main theorem.

\begin{defn}\textbf{($\ell_{2^*}$ norm of tensor column)}
Let $\tcol{x}$ be an $n_1 \times 1 \times n_3$ tensor column, we define an $\ell_{2^*}$ norm on it as follows. 
\begin{equation}
\|\tcol{x}\|_{2^*} = \sqrt{\sum\limits_{i=1}^{n_1} \sum\limits_{k=1}^{n_3} \tcol{x}_{i1k}^2 }
\end{equation}
Moreover, we have the following relationship between the $\ell_{2^*}$ norm of $\tcol{x}$ and its FFT along the third dimension $\widehat{\tcol{x}}$,
\begin{equation}
\|\tcol{x}\|_{2^*} = \frac{1}{\sqrt{n_3}} \|\widehat{\tcol{x}}\|_{2^*}
\end{equation}
where $1/n_3$ is the normalization constant.
\end{defn}

\begin{defn}\textbf{(Tensor Frobenius norm)} The induced Frobenius norm from the inner product defined above is given by,
\begin{equation}
\nonumber
\begin{aligned}
\|\T{A}\|_F &= \langle \T{A}, \T{A} \rangle^{1/2} = \frac{1}{\sqrt{n_3}} \|\widehat{\TA}\|_F = \sqrt{ \sum_{i} \sum_{j} \sum_{k} \T{A}_{ijk}^2 } 
\end{aligned}
\end{equation}
\end{defn}

\begin{defn}\textbf{(Tensor nuclear norm)}
The tensor nuclear norm of a tensor $\TA$ is defined in \cite{DBLP:journals/corr/SemerciHKM13}, denoted as $\|\T{A}\|_{TNN}$, is the sum of singular values of all the frontal slices of $\widehat{\T{A}}$, and is proved to be the tightest convex relaxation to l-1 norm of the tensor multi-rank \cite{Zhang:2014:NMM:2679600.2680311}. In fact,
\begin{equation}
\|\T{A}\|_{TNN} = \|\xbar{\T{A}}\|_*
\end{equation}
\end{defn}

\begin{defn}\textbf{(Tensor spectral norm)}
The tensor spectral norm $\|\T{A}\|$ of a third-order tensor $\T{A}$ is defined as the largest singular value of $\T{A}$. Moreover,
\begin{equation}
\|\T{A}\| = \|\xbar{\T{A}}\|
\end{equation}
i.e. the tensor sepctral norm of $\T{A}$ equals to the matrix spectral norm of $\xbar{\T{A}}$.
\end{defn}

\begin{defn}\textbf{(Tensor operator norm)}
Suppose $\mathfrak{L}$ is a tensor operator, then the operator norm of $\mathfrak{L}$ is defined as follows:
\begin{equation}
\|\mathfrak{L}\|_{op} = \sup_{\T{X}:\|\T{X}\|_F \le 1} \|\mathfrak{L}(\T{X})\|_F.
\end{equation}
which is consistent with matrix case. Spectral norm is equivalent to the operator norm if the tensor operator $\mathfrak{L}$ can be represented as a tensor $\T{L}$ t-product $\T{X}: \mathfrak{L}(\T{X}) = \T{L}*\T{X}$. Then $\|\mathfrak{L}\|_{op} = \|\T{L}\|$.
\end{defn}

\begin{defn}\textbf{(Tensor infinity norm)}
The tensor infinity norm $\|\T{A}\|_\infty$ is defined as follows:
\begin{equation}
\|\T{A}\|_\infty = \max_{i,j,k} |\T{A}_{ijk}|
\end{equation}
which is the entry with the largest absolute value of $\T{A}$.
\end{defn}

\section{Main Result}
\label{sec:3}
\subsection{Tensor Completion with Random Sampling}
In this section we will formally define the tensor completion problem with random sampling. For a given third-order tensor $\T{M} \in \mathbb{R}^{n_1 \times n_2 \times n_3}$ of tubal-rank $r$, and suppose there are $m$ entries of $\T{M}$ are sampled according to the Bernoulli model, which means each entry in the tensor is sampled with probability $p$ independent of others. The task of tensor completion problem is to recover $\T{M}$ from the observed entries. 


In this paper we follow the approach taken in \cite{Zhang:2014:NMM:2679600.2680311} which solves the following convex optimization problem for tensor completion.
\begin{equation}
\label{eq:TNN_min}
\begin{aligned}
\min_{\T{X}} \hspace{2mm}&\|\T{X}\|_{TNN}\\
\mbox{subject to  } &\T{X}_{ijk} = \T{M}_{ijk}, (i,j,k)\in \Omega
\end{aligned}
\end{equation}
where $\Omega$ is the set of observed entries. We will analyze the sufficient conditions under which the optimal solution to (\ref{eq:TNN_min}) is equal to $\T{M}$.

One definition is required before we state the main result. Similarly to the matrix completion case, recovery is hopeless if most of the entries are equal to zero \cite{exact_matrix_completion_candes_benj}. For tensor completion, the fact is that if tensor $\T{M}$ only has a few entries which are not equal to zero and let $\T{U}*\T{S}*\T{V}^{\Tra} = \T{M}$ be the reduced t-SVD of $\T{M}$, then the singular tensors $\T{U}$ and $\T{V}$ will be highly concentrated. So borrowing the idea of \cite{exact_matrix_completion_candes_benj}, the tensor columns, $\T{U}(:,i,:)$ and $\T{V}(:,i,:),i=1,2,...,r$ need to be sufficiently spread, which means it should be uncorrelated with the standard tensor basis. This idea motivates the following \emph{standard tensor incoherent condition} for tensors:

\begin{defn}(Standard Tensor Incoherent Condition)
Let the reduced t-SVD of a tensor $\T{M}$ is $\T{U} *\T{S} *\T{V}^\top$. $\T{M}$ is said to satisfy the \emph{standard tensor incoherent condition}, if there exists $\mu_0 > 0$ such that
\begin{equation}
\begin{aligned}
\label{eq:sic}
\max_{i=1,...,n_1} \left\| \TU\Tra * \tcol{e}_i \right\|_{2^*} &\le \sqrt{\frac{\mu_0 r}{n_1}}, \\
\max_{j=1,...,n_2} \left\| \TV\Tra * \tcol{e}_j \right\|_{2^*} &\le \sqrt{\frac{\mu_0 r}{n_2}},
\end{aligned}
\end{equation}
where $\tcol{e}_i$ is the $n_1 \times 1 \times n_3$ column basis with $\tcol{e}_{i11} = 1$ and $\tcol{e}_j$ is the $n_2 \times 1 \times n_3$ column basis with $\tcol{e}_{j11} = 1$ 
\end{defn}

Note that the smallest $\mu_0$ is $1$ achieved by the case when each tensor column $\tcol{u}_i = \TU(:,i,:)$ has entries with magnitude $1/\sqrt{n_1n_3}$, or each tensor column $\tcol{v}_i = \TV(:,i,:)$ has entries with magnitude $1/\sqrt{n_2n_3}$. The largest possible value of $\mu_0$ is $\mbox{min}(n_1,n_2)/r$ when one of the tensor columns of $\TU$ is equal to the standard tensor column basis $\tcol{e}_i$. With low $\mu_0$, each entry of $\T{M}$ is supposed to carry approximately same amount of information. 

In \cite{exact_matrix_completion_candes_benj} for matrix completion case, another joint incoherence condition  is needed, which is to bound the maximum entry of $UV\Tra$, where $U$ and $V$ are from the SVD of matrix $M=USV\Tra$. However, this joint incoherence condition is regarded unintuitive and restrictive. \cite{DBLP:journals/corr/Chen13e} successfully remove this joint incoherence by using the $\ell_{\infty,2}$ norm to get a similar bound in the dual certificate step. In our tensor completion case, we apply this idea on tensors and successfully avoid the joint incoherence condition.

Now we will state our main result.
\begin{theorem}
\label{thm:main}
Suppose $\T{M}$ is an $n_1 \times n_2 \times n_3$ tensor and its reduced t-SVD is given by $\T{M} = \T{U}*\T{S}*\T{V}^{\Tra}$ where $\T{U} \in \mathbb{R}^{n_1 \times r \times n_3}$, $\T{S} \in \mathbb{R}^{r \times r \times n_3}$ and $\T{V} \in \mathbb{R}^{n_2 \times r \times n_3}$. Suppose $\T{M}$ satisfies the standard tensor incoherent condition(\ref{eq:sic}) with parameter $\mu_0>0$. Then there exists constants $c_0,c_1,c_2 >0$ such that if
\begin{equation}
\label{eq:p}
p \ge c_0 \frac{\mu_0 r \log (n_3(n_1+n_2))}{\min \{n_1,n_2\}}.
\end{equation}
Then $\T{M}$ is the unique minimizer to (\ref{eq:TNN_min}) with probability at least $1-c_1((n_1+n_2)n_3)^{-c_2}$.
\end{theorem}

The sampling model we use here is the Bernoulli model. There are some other widely used models include sampling with or without replacement. For matrix completion the recovery guarantees for different models are consistent, with only a change of the constant factors factors\cite{DBLP:journals/tit/CandesT10,DBLP:journals/corr/abs-1001-2738}. We believe a similar result holds for tensors, and the analysis is in progress.

Note that we borrowed some idea of the matrix completion in our proof of \textbf{Theorem}~\ref{thm:main}. However, the technique of the proof is not as straightforward as the matrix completion case. First of all, we are sampling in the original domain, but the tensor nuclear norm (TNN) is defined in the Fourier domain. In fact, let $\T{P}_\Omega(\T{Z})$ denotes the same size tensor as $\T{Z}$ with $\T{P}_\Omega(\T{Z})_{ijk} = \T{Z}_{ijk}$ if $(i,j,k)\in\Omega$ and zero otherwise. Then (\ref{eq:TNN_min}) can be rewritten as
\begin{equation}
\nonumber
\begin{aligned}
\min_{\T{X}} \hspace{2mm}&\|\T{X}\|_{TNN}\\
\mbox{subject to  } &\T{P}(\T{X}) = \T{P}(\T{M}) \,\, .
\end{aligned}
\end{equation}
And it is equivalent to the following
\begin{equation}
\label{eq:F}
\begin{aligned}
\min_{\T{X}} \hspace{2mm}&\|\xbar{\T{X}}\|_*\\
\mbox{subject to  } &\T{F}\T{P}\T{F}^{-1}(\xbar{\T{X}}) = \T{F}\T{P}\T{F}^{-1}(\xbar{\T{M}}) \,\, ,
\end{aligned}
\end{equation}
where $\T{F}$ is a mappling which maps a third order tensor $\T{Z}$ to $\xbar{\T{Z}}$, and $\T{F}^{-1}$ is its inverse transform. Now the above problem is a matrix completion problem under linear constraint and it is completely defined in the Fourier domain. However, instead of deriving the provable recovery conditions for the random linear operator $\T{F}\T{P}\T{F}^{-1}$ slice-wise in the Fourier domain, in this paper we will closely follow the analysis in \cite{simpler_approach_benj,DBLP:journals/corr/Chen13e} and implicitly tackle the recovery guarantees for the random sampling operator. 

 
It is important to note that under the random sampling of the tensor, the problem does not reduce to slice-wise matrix completion under random sampling in the Fourier domain since the measurements are coupled. However this is indeed the case for \emph{random tubal sampling} \cite{Yanglet}, which does separate into individual matrix completion, with random sampling in the Fourier domain as we describe next. 

\subsection{Tensor Completion with Random Tubal Sampling}
Another way to sample the tensor is to perform random or adaptive tubal sampling as considered in an upcoming paper \cite{Yanglet} for fingerprinting application. Here we will comment on the random tubal sampling. Instead of randomly sample entries of a third-order tensor $\T{M}$ as above, one  can randomly sample tensor tubes along the third dimension. One can use the following optimization algorithm for recovering the tensor
\begin{equation}
\label{eq:tubal sampling}
\begin{aligned}
\min_{\T{X}} \hspace{2mm}&\|\T{X}\|_{TNN}\\
\mbox{subject to  } &\T{X}_{ijk} = \T{M}_{ijk}, (i,j)\in \Omega, k=1,2,...,n_3,
\end{aligned}
\end{equation}
where $\Omega$ is the set of observed tubes. It is easy to see that solving this optimization problem is equivalent to solving $n_3$ matrix completion problems in the Fourier domain. 
\begin{equation}
\nonumber
\begin{aligned}
\min_{\hat{\T{X}}^{(k)}} \hspace{2mm}& \|\hat{\T{X}}^{(k)}\|_*\\
\mbox{subject to  } &\hat{\T{X}}^{(k)}_{ij} = \hat{\T{M}}^{(k)}_{ij}, (i,j)\in \Omega
\end{aligned}
\end{equation}
for $k = 1,2,...,n_3$. Therefore tensor completion problem with tubal sampling is essentially the matrix completion from random samplings of each slice in the Fourier domain. Then we can directly use the result of matrix completion here. 

Suppose there are $p$ third-dimensional tubes of $\T{M} \in \mathbb{R}^{n_1 \times n_2 \times n_3}$ are sampled according to the Bernoulli model, which means each tube in the tensor is sampled with probability $p$ independent of the other tubes. Then we have the following theorem, which also appears in \cite{Yanglet}.
\begin{theorem}
\label{thm:tubal sampling}
Let $\T{M}$ be an $n_1 \times n_2 \times n_3$ tensor and its reduced t-SVD is given by $\T{M} = \T{U}*\T{S}*\T{V}^{\Tra}$ where $\T{U} \in \mathbb{R}^{n_1 \times r \times n_3}$, $\T{S} \in \mathbb{R}^{r \times r \times n_3}$ and $\T{V} \in \mathbb{R}^{n_2 \times r \times n_3}$. Suppose each frontal slice $\hat{\T{M}}^{(i)}$ satisfies the matrix weak incoherent condition (\ref{eq:matrix incoherency condition}) with parameter $\mu_0>0$. Then there exists constants $c_0,c_1,c_2 >0$ such that if
\begin{equation}
\nonumber
p \ge c_0 \frac{\mu_0 r \log^2 (n_1 + n_2)}{min\{n_1,n_2\}}
\end{equation}
Then $\T{M}$ is the unique minimizer to (\ref{eq:TNN_min}) with probability at least $1-c_1 n_3 (n_1+n_2)^{-c_2}$.
\end{theorem}

\textbf{A comment on the incoherency conditions for tubal sampling}: In Theorem~\ref{thm:tubal sampling} each slice $\hat{\T{M}}^{(k)}$ in the Fourier domain needs to satisfy \emph{matrix weak incoherence condition}\cite{DBLP:journals/corr/Chen13e} with parameter $\mu_0$. That is for all $k \in \{ 1,2,...,n_3\}$,
\begin{equation}
\label{eq:matrix incoherency condition}
\begin{aligned}
&\max_{i = 1,2,..,n_1} \|\hat{\T{U}}^{(k)\top} \V{e}_i\|_2 \le \sqrt{\frac{\mu_0 r}{n_1}}, \\
&\max_{j = 1,2,..,n_2} \|\hat{\T{V}}^{(k)\top} \V{e}_j\|_2 \le \sqrt{\frac{\mu_0 r}{n_2}}, 
\end{aligned}
\end{equation}
where $\V{e}_i$ denotes the $i$-th standard basis in $\mathbb{R}^n$. Note that similarly to our tensor case $1\le \mu_0 \le \min\{n_1,n_2\}/r$. However, the matrix incoherence condition of Equation~(\ref{eq:matrix incoherency condition}) is not equivalent to the standard tensor incoherent conditions of Equations~(\ref{eq:sic}). In fact from Equations~(\ref{eq:matrix incoherency condition}) we have

\begin{align}
\label{comment:1}
&\max_{i \in \{1,2,..,n_1\}} \|\hat{\T{U}}^{(k)\top} \V{e}_i\|_2 \le \sqrt{\frac{\mu_0 r}{n_1}} \\
\label{comment:2}
\Longrightarrow &\max_{i = 1,2,..,n_1} \sum_{k=1}^{n_3} \|\hat{\T{U}}^{(k)\top} \V{e}_i\|_2^2 \le \frac{n_3\mu_0 r}{n_1} \\
\nonumber
\Longleftrightarrow &\max_{i = 1,2,..,n_1} \|\hat{\T{U}}^{\top}(:,i,:)\|_{2^*}^2 \le \frac{n_3\mu_0 r}{n_1} \\
\nonumber
\Longleftrightarrow &\max_{i = 1,2,..,n_1} \left\| \TU\Tra * \tcol{e}_i \right\|_{2^*} \le \sqrt{\frac{\mu_0 r}{n_1}}, 
\end{align}
Similarly we can get
\begin{equation}
\nonumber
\max_{j=1,...,n_2} \left\| \TV\Tra * \tcol{e}_j \right\|_{2^*} \le \sqrt{\frac{\mu_0 r}{n_2}},
\end{equation}
\noindent which is exactly the standard tensor incoherent condition. Now we can conclude our standard tensor incoherent condition from the matrix weak incoherence condition, but not vice versa since (\ref{comment:2}) does not imply (\ref{comment:1}). So the standard tensor incoherent condition is much weaker and in the following section we will use this condition to prove our main theorem.

\section{Proof of the main results}
\label{sec:4}

In this section we prove the Theorem~\ref{thm:main}. The main idea is to use the convex analysis to verify that $\T{M}$ is the unique minimum tensor nuclear norm solution to (\ref{eq:TNN_min}).

To simplify the notation, without loss of generality, we assume $n_1 = n_2 = n$ and we don't put any assumption on $n_3$. For the case $n_1 \neq n_2$ it is proved in the exact same manner. 

Before continuing, some notations which are used in the proof need to be mentioned. Define a random variable $\delta_{ijk} = \mathbf{1}_{(i,j,k)\in \Omega}$ where $\mathbf{1}_{(\cdot)}$ is the indicator function. Define a projection $\T{R}_{\Omega}:\mathbb{R}^{n \times n \times n_3}\rightarrow \mathbb{R}^{n \times n \times n_3}$ as follows.
\begin{equation}
\T{R}_\Omega(\T{Z}) = \sum_{i,j,k} \frac{1}{p}\delta_{ijk}\T{Z}_{ijk} \tcol{e}_i * \tube{e}_k * \tcol{e}_j\Tra
\end{equation}
Define another two projections $\T{P}_{T}$  and $\T{P}_{T^\perp}$ as follows,
\begin{equation}
\nonumber
\begin{aligned}
\T{P}_{T}(\T{Z}) = &\TU*\TU\Tra*\T{Z} + \T{Z}*\T{V}*\T{V}\Tra\\
& - \TU*\TU\Tra * \T{Z} * \T{V}*\T{V}\Tra \\
\T{P}_{T^\perp}(\T{Z}) = & \T{Z} - \T{P}_{T}(\T{Z}) \\
= &(\T{I}-\TU*\TU\Tra)*\T{Z}*(\T{I}-\TV*\TV\Tra)
\end{aligned}
\end{equation}
where $\T{I}$ is the identity tensor of size $n \times n \times n_3$. Notice that for any $\T{A},\T{B} \in \mathbb{R}^{n_1 \times n_2 \times n_3}$, we have $\langle \T{P}_T(\T{A}),\T{P}_{T^\perp}(\T{B})   \rangle = 0$. It is straightforward to verify from the definition in the Fourier domain.

Similarly to the matrix completion case, we construct a \emph{tensor dual certificate} $\T{Y}$ and show it satisfies certain conditions. We define a $\ell_{\infty,2^*}$ norm for tensors. It returns the largest $\ell_{2^*}$ norm of the tensor row or tensor column of a third-order tensor.
\begin{equation}
\nonumber
\begin{aligned}
&\|\T{Z}\|_{\infty,2*}\\
:= &\max \left\{ \max_{i} \sqrt{\sum_{b,k}\T{Z}_{ibk}^2}, \max_{j} \sqrt{\sum_{a,k}\T{Z}_{ajk}^2 } \right\}
\end{aligned}
\end{equation}

The following proposition and lemma directly support the proof of the main theorem and will be proved in the appendix.
\begin{proposition}
\label{pro:1}
Suppose $p$ satisfies (\ref{eq:p}), then tensor $\T{M}$ is the unique minimizer to (\ref{eq:TNN_min}) if the following conditions hold.\\
\begin{enumerate}[1.~]
\item $\|\T{P}_{T} \T{R}_\Omega \T{P}_{T} - \T{P}_{T} \|_{op} \le \frac{1}{2}$\\
\item There exists a tensor $\T{Y}$ such that $\T{P}_\Omega(\T{Y}) = \T{Y}$ and\\
\begin{enumerate}[(a)]
\item $\|\T{P}_T(\T{Y}) - \TU * \TV\Tra \|_F \le \frac{1}{4nn_3^2}$\\
\item $\|\T{P}_{T^\perp}(\T{Y})\| \le \frac{1}{2}$
\end{enumerate}
\end{enumerate}
\end{proposition}

\begin{lemma}
\label{lemma:final}
Suppose $\|\T{P}_T \T{R}_\Omega \T{P}_T - \T{P}_T \|_{op} \le \frac{1}{2}$. Then for any $\T{Z}$ such that $\T{P}_\Omega (\T{Z}) = 0$, we have
\begin{equation}
\frac{1}{2}\|\T{P}_{\T{T}^\perp}(\T{Z})\|_{TNN} > \frac{1}{4nn_3}\|\T{P}_T(\T{Z})\|_F 
\end{equation}
with very high probability. 
\end{lemma}

\noindent \textbf{Proof of Theorem~\ref{thm:main}}
We now proceed our proof of the main theorem.
\begin{proof}
The main idea is we want to prove under these conditions, for any $\T{Z}$ supported in $\Omega^c$, $\|\T{M}+\T{Z}\|_{TNN} > \|\T{M}\|_{TNN}$. The following 3 facts are used in the main proof.
\vspace{-2mm}
\begin{fact}
\label{fact:4.1}
$\|\T{A}\|_{TNN} = n_3\sup_{\|\T{B}\|\le 1 } \langle \T{A} , \T{B}\rangle$, where $\T{A} , \T{B} \in \mathbb{R}^{n \times n \times n_3}$. Specifically, if the t-SVD of $\T{A}$ is given by $\T{A} = \T{U}*\T{S}*\T{V}^{\Tra}$, then let $\T{B} = \T{U}*\T{V}^{\Tra}$. Obviously $\|\TB\| \le 1$ and we have $n_3\langle \T{A} , \T{B}\rangle = \mbox{trace}(\xbar{\TS}) = \|\T{A}\|_{TNN}$.
\end{fact}
\vspace{-2mm}
Recall that for matrix case, we have $\|A\|_* = \sup_{\|B\|\le 1 } \langle A , B\rangle$, where $A,B \in \mathbb{R}^{n \times n}$. Then the fact is coming from
\begin{equation}
\nonumber
\begin{aligned}
\|\TA\|_{TNN} = &\|\xbar{\TA}\|_*
=  \sup_{\|\xbar{\TB}\|\le 1 } \langle \xbar{\TA} , \xbar{\TB} \rangle\\
= & n_3\sup_{\|\T{B}\|\le 1 } \langle \T{A} , \T{B}\rangle
\end{aligned}
\end{equation}

Define the t-SVD of $\T{P}_{\T{T}^\perp} (\T{Z})$ to be $\T{P}_{\T{T}^\perp} (\T{Z}) = \T{U}_\perp*\T{S}_\perp*\T{V}^{\Tra}_\perp$, where $\T{Z} \in \mathbb{R}^{n \times n \times n_3}$ such that $\T{P}_\Omega(\T{Z}) = 0$. Then use the fact above we have
\begin{equation}
\label{eq:PTperpZ_TNN}
\|\T{P}_{\T{T}^\perp} (\T{Z})\|_{TNN} = n_3\langle \T{U}_\perp*\T{V}^{\Tra}_\perp , \T{P}_{\T{T}^\perp} (\T{Z}) \rangle
\end{equation}
\vspace{-1mm}
\begin{fact} $\|\T{M}\|_{TNN} = n_3 \langle \TU*\TV\Tra + \TU_\perp * \TV\Tra_\perp , \T{M}\rangle$
\label{fact:4.2}
\end{fact}
\vspace{-1mm}
Since $\T{P}_T(\TU) = \TU$ and $\T{P}_{T^\perp}(\TU_{\perp}) = \TU_\perp$, we have $\TU*\TU\Tra_\perp = 0$ and similarly $\TV*\TV\Tra_\perp = 0$ by definition.

Then the fact can be verified by the following
\begin{equation}
\begin{aligned}
\nonumber
&n_3\langle \T{U}*\T{V}^{\Tra}+\T{U}_\perp*\T{V}^{\Tra}_\perp , \T{M} \rangle \\
= &n_3\langle \T{U}*\T{V}^{\Tra}+\T{U}_\perp*\T{V}^{\Tra}_\perp , \T{U}*\T{S}*\T{V}^{\Tra} \rangle \\
= &\text{trace}{((\xbar{\TU}\xbar{\TV}}\Tra + \xbar{\TU}_\perp \xbar{\TV}\Tra_\perp )\Tra \hspace{1mm} \xbar{\TU} \xbar{\TS} \xbar{\TV}\Tra  ) \\
= &\text{trace}(\xbar{\T{S}} ) = \|\T{M}\|_{TNN}
\end{aligned}
\end{equation}
\begin{fact}$\|\TU*\TV\Tra + \TU_\perp * \TV\Tra_\perp  \| = 1$
\label{fact:4.3}
\end{fact}
Consider a matrix $Q$ such that
\begin{equation}
\nonumber
Q = \xbar{\T{U}}\xbar{\T{V}}\Tra + \xbar{\T{U}}_\perp \xbar{\T{V}}\Tra_\perp =
  \left[\begin{array}{ccc}\xbar{\T{U}} &\xbar{\T{U}}_\perp  \end{array}\right] \cdot \left[\begin{array}{c}\xbar{\T{V}}\Tra\\ \xbar{\T{V}}\Tra_\perp  \end{array}\right] 
\end{equation}
Since $\xbar{\T{U}}\Tra \xbar{\T{U}}_\perp = 0$ and $\xbar{\T{V}}\Tra \xbar{\T{V}}_\perp=0$, the above expression is the matrix singular value decomposition of $Q$, so we have
\begin{equation}
\nonumber
\begin{aligned}
\|\T{U}*\T{V}^{\Tra} + \T{U}_\perp * \T{V}^{\Tra}_\perp \| 
&= \| \xbar{\T{U}}\xbar{\T{V}}\Tra + \xbar{\T{U}}_\perp \xbar{\T{V}}\Tra_\perp \| \\
&= \|Q\| = 1
\end{aligned}
\end{equation}

Now using the above facts, given any $\T{Z} \in \mathbb{R}^{n \times n \times n_3}$ such that $\T{P}_\Omega(\T{Z}) = 0$, we have
\begin{align}
\nonumber
&\|\T{M}+\T{Z}\|_{TNN}\\
\label{eq:fact1}
\ge & n_3 \langle \T{U}*\T{V}^{\Tra} + \T{U}_\perp * \T{V}^{\Tra}_\perp , \T{M}+\T{Z} \rangle \\
\nonumber 
=   & \|\T{M}\|_{TNN} + n_3 \langle \T{U}*\T{V}^{\Tra} + \T{U}_\perp * \T{V}^{\Tra}_\perp , \T{Z} \rangle \\
\nonumber
=   & \|\T{M}\|_{TNN} + n_3 \langle \T{U}*\T{V}^{\Tra} , \T{P}_{T}(\T{Z})\rangle \\
\nonumber
    & + n_3 \langle  \T{U}_\perp * \T{V}^{\Tra}_\perp, \T{P}_{T^\perp}(\T{Z}) \rangle \\
\nonumber
=   & \|\T{M}\|_{TNN} + n_3 \langle \T{U}*\T{V}^{\Tra} , \T{P}_{T}(\T{Z})\rangle \\
\label{eq:main_1}	
    & + n_3\langle  \T{U}_\perp * \T{V}^{\Tra}_\perp, \T{P}_{T^\perp}(\T{Z}) \rangle  - n_3\langle \T{Y} , \T{Z} \rangle \\
\nonumber
=   & \|\T{M}\|_{TNN} + n_3 \langle \T{U}*\T{V}^{\Tra} - \T{P}_{T}(\T{Y}) , \T{P}_{T}(\T{Z})\rangle  \\
\nonumber
    & + n_3\langle  \T{U}_\perp * \T{V}^{\Tra}_\perp - \T{P}_{T^\perp}(\T{Y}), \T{P}_{T^\perp}(\T{Z}) \rangle \\
\nonumber    
=   & \|\T{M}\|_{TNN} + \langle \xbar{\TU}\xbar{\TV}\Tra - \xbar{\T{P}_{T}(\T{Y})} , \xbar{ \T{P}_{T}(\T{Z})} \rangle  \\
\label{eq:main_2}
    & +  \|\T{P}_{T^\perp}(\T{Z})\|_{TNN} - \langle  \xbar{\T{P}_{T^\perp}(\T{Y})} , \xbar{\T{P}_{T^\perp}(\T{Z})} \rangle \\
\nonumber
\ge & \|\T{M}\|_{TNN} - \|\xbar{\TU}\xbar{\TV}\Tra - \xbar{\T{P}_{T}(\T{Y})}\|_F  \|\xbar{ \T{P}_{T}(\T{Z})}\|_F \\
\label{eq:main_3}
   & +  \|\T{P}_{T^\perp}(\T{Z})\|_{TNN} - \|\xbar{\T{P}_{T^\perp}(\T{Y})}\| \|\xbar{\T{P}_{T^\perp}(\T{Z})}\|_* \\
\nonumber
= & \|\T{M}\|_{TNN} - n_3 \|\TU*\TV\Tra - \T{P}_{T}(\T{Y})\|_F  \|\T{P}_{T}(\T{Z})\|_F \\
\nonumber
   & + \|\T{P}_{T^\perp}(\T{Z})\|_{TNN} - \|\T{P}_{T^\perp}(\T{Y})\| \|\T{P}_{T^\perp}(\T{Z})\|_{TNN} \\
\label{eq:main_4}
\ge & \|\T{M}\|_{TNN} - \frac{1}{4nn_3}\|\T{P}_{T}(\T{Z})\|_F + \frac{1}{2}\|\T{P}_{T^\perp}(\T{Z})\|_{TNN} \\
\nonumber
> & \|\T{M}\|_{TNN}
\end{align}
\noindent where (\ref{eq:fact1}) uses the Fact~\ref{fact:4.1}; $\T{Y}$ in (\ref{eq:main_1}) is a tensor dual certificate supported in $\Omega$ such that $\T{P}_\Omega (\T{Y}) = \T{Y}$. So it is easy to show $\langle \T{Z},\T{Y}\rangle = 0$ using the standard basis decomposition; (\ref{eq:main_2}) uses equation (\ref{eq:PTperpZ_TNN}) ; And (\ref{eq:main_3}) is based on the following two facts for any same size matrices $A$ and $B$:
\begin{align}
\nonumber
|\langle A , B\rangle | &\le \|A\|_F \|B\|_F \\
\nonumber
\langle A , B\rangle &\le \|A\| \|B\|_*
\end{align}
\noindent and (\ref{eq:main_4}) uses the Condition \textbf{2} of Proposition~\ref{pro:1}.

 Therefore, for any $\T{X} \neq \T{M}$ obeying $\T{P}_\Omega(\T{X}-\T{M})=0$, we have $\|\T{X}\|_{TNN} > \|\T{M}\|_{TNN}$, hence $\T{M}$ is the unique minimizer of (\ref{eq:TNN_min}), end of proof.

\end{proof}

\section{Numerical Experiments}
\label{sec:5}
To demonstrate our results, we conducted some numerical experiments to recover a $3$-rd order tensor of different size $n_1 \times n_2 \times n_3$ and tubal-ranks $r$, from $m$ observed entries. For each episode we generated a $n_1 \times n_2 \times n_3$ random tensor with i.i.d. Gaussian entries, performed a t-SVD of it, kept the first $r$ singular tubes and got $\T{M}$. We sampled $m$ entries of $\T{M}$ uniformly at random and try to recover $\T{M}$ using \ref{eq:TNN_min}. We call the solution  $\T{X}_0$. If the relative square error (RSE): $\|\T{X}_0 - \T{M}\|_F/\|\T{M}\|_F \le 10^{-3}$, then we say the recovery is correct. We repeated our experiments 20 times and the results are shown in Figure~\ref{fig:recovery_curve}. In the left figure, the color of each cell reflects the empirical recovery rate ranging from $0$ to $1$. White cell means exact recovery in all the experiments and black cell means all the experiments failed. The right figure is the RSE plot of one typical run of the simulation. For each cell the value reflects the RSE of the recovery under the corresponding sampling rate and tubal rank.Black denotes 1 and white denotes 0.
\begin{figure}[htbp]
	\centering \makebox[0in]{
		\begin{tabular}{c c}
			\includegraphics[scale=0.26]{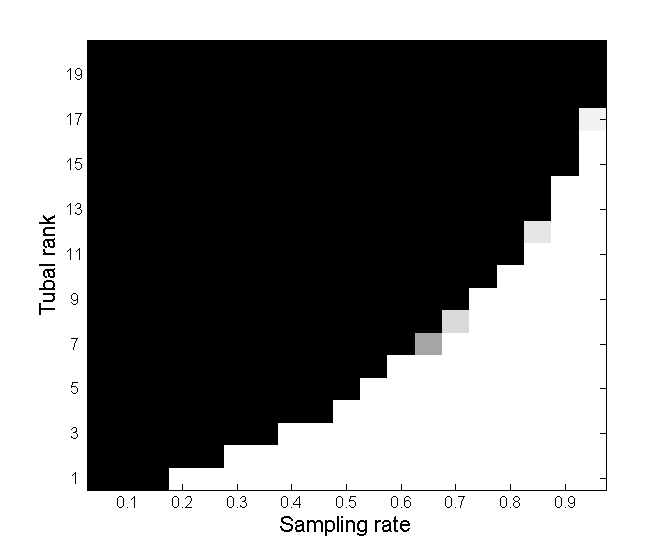} 
			\includegraphics[scale=0.26]{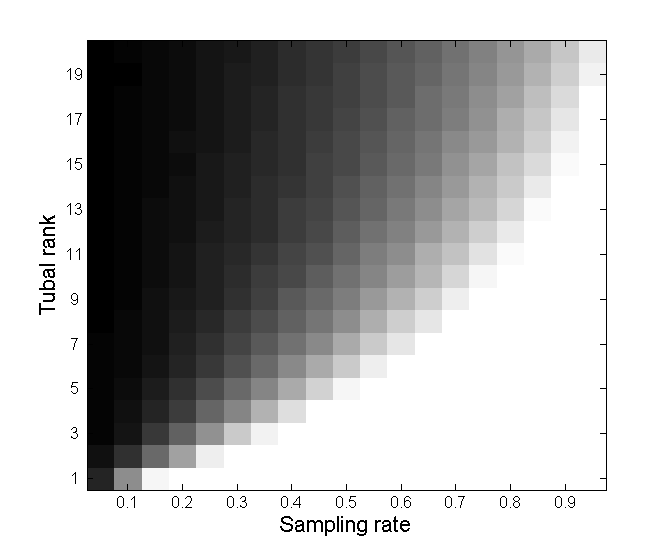}  \\
			(a)\\
			\includegraphics[scale=0.26]{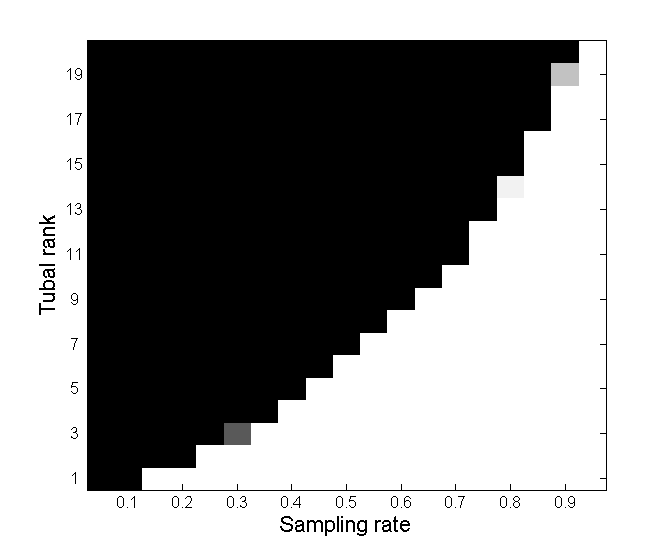} 
			\includegraphics[scale=0.26]{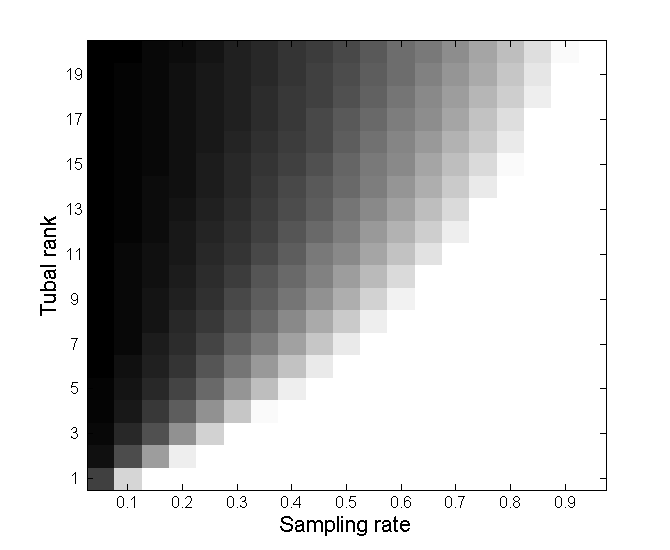}\\
		    (b)\\
		\end{tabular}}
		\caption{ (a) $40 \times 40 \times 30$ tensor. (b) $30 \times 30 \times 20$ tensor. In the left figures of both cases, each cell's value reflects the empirical recovery rate. Black denotes failure and white denotes success in recovery in all simulations. In the right figures of both cases, each cell's value is the RSE of the recovery under the corresponding sampling rate and tubal rank. Black denotes 1 and white denotes 0.}
		\label{fig:recovery_curve}
	\end{figure}
	
For practical application we tested 5 algorithms for video data completion from randomly missing entries. The first method is performing matrix completion\cite{Cai:2010:SVT:1898437.1898451} on each frame of the tensor, the second approach is the Low Rank Tensor Completion(LRTC) algorithm in \cite{6138863}, the third one is the tensor completion in Tucker format by Riemannian optimization \cite{KressSV:2014}, the fourth approach is our tensor completion with random sampling via t-SVD, and the last approach is tensor completion with tubal sampling\cite{Yanglet} as explained in \textbf{Section}~\ref{sec:3}. The size of the video is $144 \times 256 \times 80$ and we randomly sampled $50\%$ entries from the video. Note that for the first 4 approaches we used the same sampling model, but for the last approach, which is tensor completion with tubal sampling, we sampled $50\%$ tubes of the original video. The result is shown in Figure~\ref{fig:completion_result}. We compared relative square error (RSE) of each approach in our simulation and the result is in Table~\ref{table:2}, which shows that both our approach yields a better performance over the other 4 methods in the experiment.

\begin{figure}[htbp]
	\centering \makebox[0in]{
		\begin{tabular}{c c}
			\includegraphics[scale=0.3]{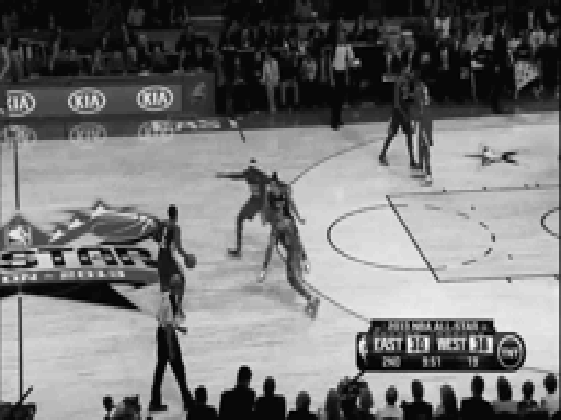} 
			\includegraphics[scale=0.3]{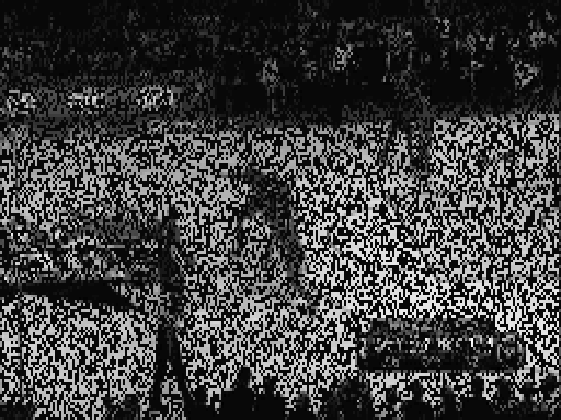} \\
			(a)\\
			\includegraphics[scale=0.3]{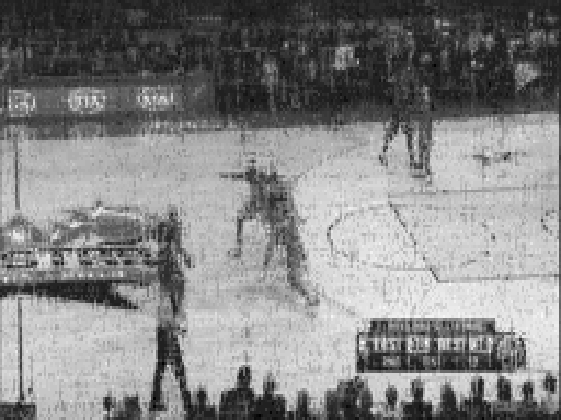} 
			\includegraphics[scale=0.3]{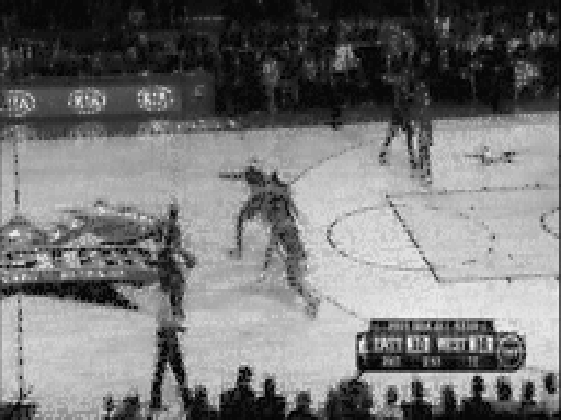} \\
			(b)\\
			\includegraphics[scale=0.3]{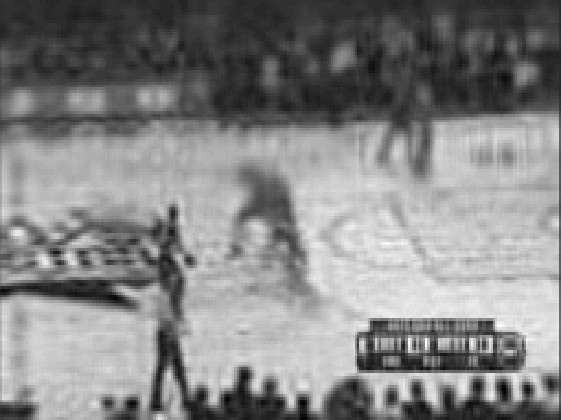} 
			\includegraphics[scale=0.3]{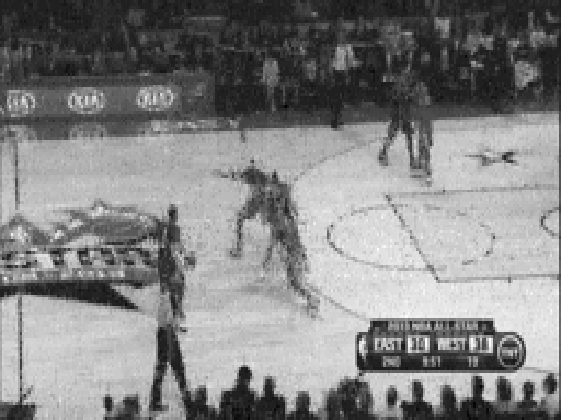}  \\
			(c)\\
			\includegraphics[scale=0.3]{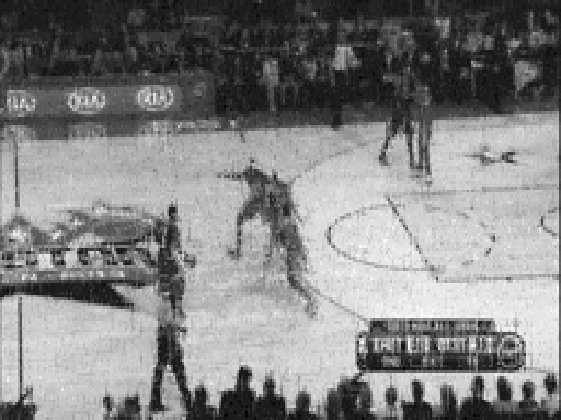} \\
			(d)\\
		\end{tabular}}
		\caption{ The $50$th frame of tensor completion result on a basketball video. \textbf{(a)Left }: The original video. \textbf{(a)Right}: Sampled video (50\% sampling rate).  \textbf{(b)Left}: Matrix completion on each frame. \textbf{(b)Right}: LRTC method. \textbf{(c)Left}: Tensor completion by Riemannian optimization using Tucker decomposition. \textbf{(c)Right}: Tensor completion with random sampling using t-SVD. \textbf{(d)}: Tensor completion with tubal sampling using t-SVD.   }
		\label{fig:completion_result}
\end{figure}
	
\begin{table}
	\begin{center}
		\begin{tabular}{|p{5cm}|p{1cm}|}
			\hline
			\textbf{Completion Approach}     & \textbf{RSE}    \\ \hline
			Matrix completion on each frame  & 0.1409 \\ \hline
			LRTC                             & 0.1218 \\ \hline
			Tensor completion in Tucker format by Riemannian optimization  & 0.1502 \\ \hline
			Tensor completion with random sampling via tSVD       & 0.0915 \\ \hline
			Tensor completion with tubal sampling via tSVD       & 0.1065 \\ \hline			
		\end{tabular}
	\end{center}
	\caption{RSE of of tensor completion result on the basketball video}
	\label{table:2}
\end{table}
\section{Conclusion}
\label{sec:6}

In this paper we have shown that under the certain Tensor Standard Incoherent condition, one can perfectly recover a tensor with low tubal-rank, and establish a theoretical bound for exact completion. We compared different tensor completion approaches in experiment and showed that our method yields a better performance over the others on some real data sets.

Our results indicate that for tensors of moderate order, t-SVD can provide a computationally as well as theoretically sound analysis framework. Moreover, in this framework the tools and methods developed to deal with prediction and learning problems for matrices using vector space methods can be adapted to obtain reliable performance guarantees.

\pagestyle{empty}
\bibliographystyle{IEEEtran}
\bibliography{bib_PAMI}



\appendices
\section{Proof of Proposition~\ref{pro:1} Condition 1}

The following theorem is first developed in \cite{simpler_approach_benj}, and will be used frequently in the following proof.

\begin{theorem}
\label{thm:NBI}
\textbf{(Noncommutative Bernstein Inequality)}\\
Let $X_1,X_2,...,X_L$ be independent zero-mean random matrices of dimension $d_1 \times d_2$. Suppose 
\begin{equation}
\nonumber
\rho^2_k = \max \{ \| \mathbb{E}[X_k X_k\Tra] \|,\|\mathbb{E}[X_k\Tra X_k]\| \}
\end{equation}
and 
\begin{equation}
\nonumber
\|X_k\| \le M
\end{equation}
almost surely for all $k$. Then for any $\tau > 0$,
\begin{equation}
\label{eq:NBI_1}
\begin{aligned}
&\mathbb{P}\left[ \left\|\sum_{k=1}^{L}X_k\right\| > \tau \right] \\
\le &(d_1+d_2)\exp \left( \frac{-\tau^2/2}{ \sum_{k=1}^{L}\rho^2_k + M\tau/3} \right)
\end{aligned}
\end{equation}
\end{theorem}
This theorem is a corollary of an Chernoff bound for finite dimension operators developed by \cite{strong_converse_A_W}. An extension of this theorem\cite{Tropp:2012:UTB:2347803.2347804} states that if
\begin{equation}
\max \left\{ \left\| \sum_{k=1}^{L}X_k X_k\Tra  \right\| , \left\| \sum_{k=1}^{L}X_k\Tra X_k  \right\|  \right\} \le \sigma^2
\end{equation}
\noindent and let
\begin{equation}
\nonumber
\tau = \sqrt{4c\sigma^2 \log(d_1+d_2)}+cM\log(d_1+d_2)
\end{equation}
for any $c>0$. Then (\ref{eq:NBI_1}) becomes 
\begin{equation}
\label{eq:NBI_2}
\mathbb{P}\left[ \left\|\sum_{k=1}^{L}X_k\right\| \ge \tau \right] \le (d_1+d_2)^{-(c-1)}
\end{equation}


The following fact is very useful and will be used frequently in this section.
\begin{fact}
$ \left \|\T{P}_{T}(\tcol{e}_i * \tube{e}_k * \tcol{e}_j\Tra ) \right\|^2_F \le \frac{2\mu_0r}{n}$
\end{fact}

\noindent \textbf{Proof of Proposition}~\ref{pro:1} \textbf{Condition} (1).

\begin{proof}
First note that 
\begin{equation}
\label{eq:temp1}
\mathbb{E}[ \T{P}_T \T{R}_\Omega \T{P}_T ] = \T{P}_T ( \mathbb{E}\T{R}_\Omega) \T{P}_T = \T{P}_T \,\, ,
\end{equation}
which gives
\begin{equation}
\nonumber
\mathbb{E}[\T{P}_T \T{R}_\Omega \T{P}_T - \T{P}_T] = 0
\end{equation}
and
\begin{equation}
\mathbb{E}[\xbar{\T{P}_T} \xbar{\T{R}_\Omega} \xbar{\T{P}_T} - \xbar{\T{P}_T}] = 0
\end{equation}

Our goal is to prove the operator $\xbar{\T{P}_T} \xbar{\T{R}_\Omega} \xbar{\T{P}_T}$ is not far away from its expected value $\xbar{\T{P}_T}$ in the spectral norm using the Noncummutative Bernstein Inequality. 

Given any tensor $\T{Z}$ of size $n \times n \times n_3$, we can decompose $\T{P}_T(\T{Z})$ as the following
\begin{equation}
\nonumber
\begin{aligned}
\T{P}_T(\T{Z}) = &\sum\limits_{i,j,k} \langle \T{P}_T(\T{Z}) , \tcol{e}_i*\tube{e}_k*\tcol{e}_j\Tra \rangle \tcol{e}_i*\tube{e}_k*\tcol{e}_j\Tra \\
= &\sum\limits_{i,j,k} \langle \T{Z} , \T{P}_T (\tcol{e}_i*\tube{e}_k*\tcol{e}_j\Tra) \rangle \tcol{e}_i*\tube{e}_k*\tcol{e}_j\Tra 
\end{aligned}
\end{equation}
This gives
\begin{equation}
\nonumber
\T{R}_\Omega \T{P}_T (\T{Z}) = \sum\limits_{i,j,k} \frac{1}{p}\delta_{ijk} \langle \T{Z} , \T{P}_T (\tcol{e}_i*\tube{e}_k*\tcol{e}_j\Tra) \rangle \tcol{e}_i*\tube{e}_k*\tcol{e}_j\Tra 
\end{equation}
\noindent and
\begin{equation}
\nonumber
\begin{aligned}
&\T{P}_T \T{R}_\Omega \T{P}_T (\T{Z}) \\
= &\sum\limits_{i,j,k} \frac{1}{p}\delta_{ijk} \langle \T{Z} , \T{P}_T (\tcol{e}_i*\tube{e}_k*\tcol{e}_j\Tra) \rangle \T{P}_T (\tcol{e}_i*\tube{e}_k*\tcol{e}_j\Tra )
\end{aligned}
\end{equation}
\noindent and this implies
\begin{equation}
\nonumber
\begin{aligned}
& \xbar{\T{P}_T} \xbar{\T{R}_\Omega} \xbar{\T{P}_T} (\xbar{\T{Z}} )  \\
= &\sum\limits_{i,j,k} \frac{1}{p}\delta_{ijk} \langle \T{Z} , \T{P}_T (\tcol{e}_i*\tube{e}_k*\tcol{e}_j\Tra) \rangle \xbar{ \T{P}_T} ( \xbar{\tcol{e}_i \tube{e}_k \tcol{e}_j\Tra } ) 
\end{aligned}
\end{equation}
Define the operator $\T{T}_{ijk}$ which maps $\T{Z}$ to $\frac{1}{p}\delta_{ijk} \langle \T{Z} , \T{P}_T (\tcol{e}_i*\tube{e}_k*\tcol{e}_j\Tra) \rangle \T{P}_T (\tcol{e}_i*\tube{e}_k*\tcol{e}_j\Tra ) $. Observe that $
\|\T{T}_{ijk}\|_{op}  = \|\xbar{\T{T}_{ijk}}\|= \frac{1}{p}\|\T{P}_T (\tcol{e}_i*\tube{e}_k*\tcol{e}_j\Tra )\|_F^2$ and $\|\T{P}_T\|_{op} = \|\xbar{\T{P}_T}\| \le 1$. Then we have
\begin{equation}
\begin{aligned}
\nonumber
&\|\T{T}_{ijk} - \frac{1}{n^2 n_3}\T{P}_T \|_{op}  \\
=& \|\xbar{\T{T}_{ijk}} - \frac{1}{n^2 n_3}\xbar{\T{P}_T} \| \\
\le &\max \left\{ \frac{1}{p}\|\T{P}_T (\tcol{e}_i*\tube{e}_k*\tcol{e}_j\Tra )\|_F^2, \frac{1}{n^2 n_3} \right\} \\
\le & \frac{2\mu_0r}{np}
\end{aligned}
\end{equation}
where the first inequality uses the fact that if $A$ and $B$ are positive semidefinite matrices, then $\|A-B\| \le \max\{\|A\|,\|B\|\}$.

On the other hand, observe that from (\ref{eq:temp1}) we have $\mathbb{E}[\T{T}_{ijk}] = \frac{1}{n^2n_3}\T{P}_T$. So
\begin{equation}
\begin{aligned}
\nonumber
&\left\|\mathbb{E}[(\xbar{\T{T}_{ijk}} - \frac{1}{n^2 n_3} \xbar{\T{P}}_T)^2]\right\| \\
=&\left\|\mathbb{E}[(\T{T}_{ijk} - \frac{1}{n^2 n_3} \T{P}_T)^2] \right\|\\
\le &\Bigg\|\mathbb{E}[\frac{1}{p}\|\T{P}_T(\tcol{e}_i*\tube{e}_k*\tcol{e}_j\Tra )\|_F^2 \T{T}_{ijk} ] - \frac{2}{n^2 n_3} \T{P}_T \mathbb{E}[\T{T}_{ijk}] \\
&+ \frac{1}{n^4 n_3^2} \T{P}_T \Bigg\| \\
= &\left\|\frac{1}{p} \|\T{P}_T(\tcol{e}_i*\tube{e}_k*\tcol{e}_j\Tra )\|_F^2 \frac{1}{n^2n_3}\T{P}_T  - \frac{1}{n^4 n_3^2} \T{P}_T\right\| \\
< &\left( \frac{1}{p} \frac{2\mu_0 r }{n} \frac{1}{n^2n_3} \right)\|\T{P}_T\| \\
\le & \frac{2\mu_0r}{n^3 n_3 p}
\end{aligned}
\end{equation}
Now let 
\begin{equation}
\nonumber
\tau = \sqrt{ \frac{14\beta\mu_0 r \log (nn_3)}{3np} } \le \frac{1}{2}
\end{equation}
\noindent with some constant $\beta>1$. The inequality holds given $p$ satisfying (\ref{eq:p}) with $c_0$ large enough. Use \textbf{Theorem} \ref{thm:NBI} we have,
\begin{equation}
\nonumber
\begin{aligned}
&\mathbb{P}\left[ \| \T{P}_T \T{R}_\Omega \T{P}_T - \T{P}_T \|_{op} > \tau \right] \\
=& \mathbb{P}\left[  \| \xbar{\T{P}_T} \xbar{\T{R}_\Omega}  \xbar{\T{P}_T} - \xbar{\T{P}_T} \| > \tau \right] \\
=& \mathbb{P}\left[ \left\| \sum_{i,j,k} \left( \xbar{\T{T}_{ijk}} - \frac{1}{n^2n_3}\xbar{\T{P}_T} \right) \right\| > \tau \right]\\
\le & 2nn_3 \exp \left( \frac{-\frac{7}{3}\frac{\beta \mu_0 r \log (nn_3)}{np} }{\frac{2\mu_0 r}{np} + \frac{2\mu_0 r}{np} \frac{1}{6} } \right) \\
\le & 2(nn_3)^{1-\beta}
\end{aligned}
\end{equation}
Then we have 
\begin{equation}
\nonumber
\begin{aligned}
&\mathbb{P}\left[ \| \T{P}_T \T{R}_\Omega \T{P}_T - \T{P}_T \|_{op} \le \frac{1}{2} \right] \\
\ge & \mathbb{P}\left[ \| \T{P}_T \T{R}_\Omega \T{P}_T - \T{P}_T \|_{op} \le \tau \right]  \\
\ge & 1-2(nn_3)^{1-\beta} \,\, ,
\end{aligned}
\end{equation}
which finishes the proof.
\end{proof}

\section{Proof of Lemma~\ref{lemma:final} }

\begin{proof}
Given any $\T{Z}$ such that $\T{P}_\Omega(\T{Z}) = 0$ and $\|\T{P}_T \T{R}_\Omega \T{P}_T - \T{P}_T\|_{op} \ge 1/2$, we have
\begin{equation}
\nonumber
\langle \xbar{\T{Z}} , \overline{\T{P}_T \T{R}_\Omega \T{P}_T(\T{Z})} -\xbar{\T{P}_T(\T{Z}) }  \rangle \ge -\frac{1}{2} \|\xbar{\T{Z}}\|_F
\end{equation}
which gives

\begin{equation}
\nonumber
\langle \T{Z} , \T{P}_T \T{R}_\Omega \T{P}_T (\T{Z}) - \T{P}_T(\T{Z})  \rangle \ge -\frac{1}{2}\|\T{Z}\|_F
\end{equation}

Note that 
\begin{equation}
\nonumber
\begin{aligned}
&\langle \T{Z} , \T{P}_T \T{R}_\Omega \T{P}_T (\T{Z})\rangle \\
= &\frac{1}{\sqrt{n_3}} \langle \xbar{\T{Z}} , \overline{\T{P}_T \T{R}_\Omega \T{P}_T (\T{Z})}\rangle \\
= &\frac{1}{\sqrt{n_3}} \| \overline{ \T{R}_\Omega \T{P}_T (\T{Z}) } \|_F^2 \\
= & \sqrt{n_3} \|\T{R}_\Omega \T{P}_T (\T{Z})\|^2_F \\
= & \sqrt{n_3} \|\T{R}_\Omega ( \T{Z} - \T{P}_{T^\perp} (\T{Z}))\|^2_F \\
= & \sqrt{n_3} \|\T{R}_\Omega \T{P}_{T^\perp} (\T{Z})\|^2_F \\
\le & \frac{\sqrt{n_3}}{p^2} \|\T{P}_{T^\perp} (\T{Z})\|^2_F
\end{aligned}
\end{equation}

Thus 
\begin{equation}
\nonumber
\begin{aligned}
&\|\T{P}_{T^\perp}(\T{Z})\|^2_F \\
\ge &\frac{p^2}{\sqrt{n_3}}\langle \T{Z} , \T{P}_T \T{R}_\Omega \T{P}_T (\T{Z})\rangle  \\
\ge &\frac{p^2}{\sqrt{n_3}}\left( -\frac{1}{2}\|\T{Z}\|_F + \langle \T{Z} , \T{P}_T(\T{Z})\rangle  \right)\\
= &\frac{p^2}{\sqrt{n_3}} \left( \frac{1}{\sqrt{n_3}} \langle \xbar{\T{Z}}, \overline{\T{P}_T(\T{Z})}\rangle - \frac{1}{2}\|\T{Z}\|_F  \right) \\
= &\frac{p^2}{n_3} \|\overline{\T{P}_T(\T{Z})}\|^2_F - \frac{p^2}{2\sqrt{n_3}} \|\T{Z}\|_F \\
\ge & (p^2 - \frac{p^2}{2\sqrt{n_3}}) \|\T{P}_T(\T{Z})\|^2_F - \frac{p^2}{2\sqrt{n_3}}\|\T{P}_{T^\perp}(\T{Z})\|^2_F
\end{aligned}
\end{equation}
Then we have
\begin{equation}
\nonumber
\begin{aligned}
\|\T{P}_{T^\perp}(\T{Z})\|^2_F \ge &p^2 \frac{2\sqrt{n_3}-1}{2\sqrt{n_3}+p^2} \|\T{P}_T(\T{Z})\|^2_F \\
\ge &\frac{1}{4n^2n_3^3}  \|\T{P}_T(\T{Z})\|^2_F 
\end{aligned}
\end{equation}

It follows that
\begin{equation}
\nonumber
\begin{aligned}
\|\T{P}_{T^\perp}(\T{Z})\|_{TNN} 
= &\|\overline{ \T{P}_{T^\perp}(\T{Z}) }\|_*\\
\ge &\|\overline{ \T{P}_{T^\perp}(\T{Z}) }\|_F \\
\ge &\sqrt{n_3} \| \T{P}_{T^\perp}(\T{Z}) \|_F \\
\ge &\frac{1}{2nn_3} \| \T{P}_{T^\perp}(\T{Z}) \|_F
\end{aligned}
\end{equation}
which finishes the proof.
\end{proof}


\section{Proof of Proposition~\ref{pro:1} Condition 2}

To finish the proof some more lemmas are needed. \textbf{NOTE}: The proofs of these Lemmas are provided in the \textbf{Supplementary Materials} for review purposes.

The following lemma states that $\T{R}_\Omega(\T{Z})$ is closed to $\T{Z}$ in tensor spectral norm. And the difference is bounded using both $\ell_\infty$ norm and the $\ell_{\infty,2^*}$ norm.
\begin{lemma}
	\label{lemma:2}
	If $p$ satisfies the condition in Theorem~\ref{thm:main}, and $\T{Z}\in \mathbb{R}^{n \times n \times n_3}$. Then for any constant $c>0$, we have
	\begin{equation}
		\begin{aligned}
			&\|\T{R}_\Omega (\T{Z}) - \T{Z}\|  \\
			\le &c\left( \frac{\log (nn_3)}{p} \|\T{Z}\|_\infty + \sqrt{\frac{\log (nn_3)}{p}}\|\T{Z}\|_{\infty,2^*} \right)
		\end{aligned}
	\end{equation}
	holds with probability at least $1-(2nn_3)^{-(c-1)}$.
\end{lemma}

The lemma below bounds the $\ell_{\infty,2^*}$ distance between $\T{P}_T\T{R}_\Omega(\T{Z})$ and $\T{P}_T(\T{Z})$.

\begin{lemma}
	\label{lemma:3}
	If $p$ satisfies the condition in Theorem~\ref{thm:main} for some $c_2$ sufficiently large, and $\T{Z}\in \mathbb{R}^{n \times n \times n_3}$. Then 
	\begin{equation}
		\nonumber
		\left\| (\T{P}_T\T{R}_{\Omega}(\T{Z}) - \T{P}_T(\T{Z}) \right\|_{\infty,2^*} 
		\le  \frac{1}{2}\|\T{Z}\|_{\infty,2^*}+ \frac{1}{2}\sqrt{\frac{n}{\mu_0 r}} \|\T{Z}\|_{\infty}
	\end{equation}
	with probability at least $1-(2n^2n_3)^{-(c_2-1)}$.
\end{lemma}

Using \textbf{Theorem}~\ref{thm:NBI}, we have the following Lemma.
\begin{lemma}
	\label{lemma:4}
	If $p$ satisfies the condition in Theorem~\ref{thm:main} for some $c_3$ sufficiently large, and $\T{Z}\in \mathbb{R}^{n \times n \times n_3}$. Then 
	\begin{equation}
		\| (\T{P}_T\T{R}_{\Omega} \T{P}_T - \T{P}_T ) \T{Z} \|_\infty \le \frac{1}{2} \|\T{Z}\|_\infty
	\end{equation}
	
	with probability at least $1-2n^{-(c_3-2)}n_3^{-(c_3-1)}$.
\end{lemma}

Using these Lemmas we can now prove the Proposition~\ref{pro:1} Condition 2.\\
\begin{proof}
\textbf{(a)} We will first construct a tensor helper $\T{Y}$ and then show it satisfies both conditions here. We will use an approach called Golfing Scheme introduced by Gross \cite{DBLP:journals/tit/Gross11} and we will follow the idea in \cite{simpler_approach_benj}\cite{DBLP:journals/corr/Chen13e} where the strategy is to construct $\T{Y}$ iteratively. Let $\Omega$ be a union of smaller sets $\Omega_t$ such that $\Omega = \cup_{t=1}^{t_0}$ where $t_0 = 20 \log(nn_3)$. For each $t$, we assume
\begin{equation}
\nonumber
\mathbb{P}\left[ (i,j,k) \in \Omega_t  \right] = q :=1-(1-p)^{1/t_0}
\end{equation}
and it is easy to verify that it's equivalent to our original $\Omega$. Define $\T{R}_{\Omega_t}$ similarly to $\T{R}_\Omega$ as follows
\begin{equation}
\nonumber
\T{R}_{\Omega_t} (\T{Z}) = \sum_{i,j,k} \frac{1}{q}\mathbf{1}_{(i,j,k)\in \Omega_t} \T{Z}_{ijk}\tcol{e}_i * \tube{e}_k * \tcol{e}_j\Tra  
\end{equation}
set $\T{W}_0 = 0$ and for $t=1,2,...,t_0$,
\begin{equation}
\label{eq:prop1}
\T{W}_t = \T{W}_{t-1} + \T{R}_{\Omega_t}\T{P}_T \left( \TU*\TV\Tra -\T{P}_T(\T{W}_{t-1}) \right)
\end{equation}
and the helper tensor $\T{Y} = \T{W}_{t_0}$. By this construction we can see $\T{P}_{\Omega}(\T{Y}) = \T{Y}$.

For $t= 0,1,...,t_0$, set $\T{D}_t = \TU * \TV\Tra - \T{P}_T(\T{W}_{t})$. Then we have $\T{D}_0 = \TU*\TV\Tra$ and 
\begin{equation}
\label{eq:pro_tmp_1}
\T{D}_t = (\T{P}_T - \T{P}_T \T{R}_{\Omega_t} \T{P}_T )( \T{D}_{t-1} )
\end{equation}

Note that $\Omega_t$ is independent of $\T{D}_t$, we have 
\begin{equation}
\nonumber
\left\| \T{D}_t \right\|_F \le \left\| \T{P}_T - \T{P}_T \T{R}_{\Omega_t} \T{P}_T  \right\| \left\|\T{D}_{t-1} \right\| \le \frac{1}{2} \| \T{D}_{t-1} \|_F
\end{equation}
since $q \ge p/t_0 \ge c' \mu_0 r \log(nn_3)/n$, from \textbf{Proposition}~\ref{pro:1} \textbf{condition (1)} we have
\begin{equation}
\nonumber
\begin{aligned}
&\left\| \T{P}_T(\T{Y}) - \TU*\TV\Tra \right\|_F \\
= &\left\| \T{D}_{t_0}\right\| \\
\le &\left( \frac{1}{2} \right)^{t_0} \left\| \TU*\TV\Tra \right\|_F \\
\le &\frac{1}{4(nn_3)^2}\sqrt{r} \\
\le &\frac{1}{4nn^2_3}
\end{aligned}
\end{equation}

\textbf{(b)} From (\ref{eq:prop1}) we know that $\T{Y} = \T{W}_{t_0} = \sum_{t=1}^{t_0} \left( \T{R}_{\Omega_t} \T{P}_T \right) (\T{D}_{t-1})$, so use \textbf{Lemma}~\ref{lemma:2} we obtain for some constant $c>0$,
\begin{equation}
\nonumber
\begin{aligned}
&\left\| \T{P}_{T^\perp} (\T{Y})  \right\| 
\le \sum_{t=1}^{t_0} \left\| \T{P}_{T^\perp}\left(\T{R}_{\Omega_t} \T{P}_T \right)(\T{D}_{t-1}) \right\| \\
\le &\sum_{t=1}^{t_0}\left\| \left( \T{R}_{\Omega_t} - \T{I} \right) \T{P}_T (\T{D}_{t-1})  \right\| \\
\le &c \sum_{t=1}^{t_0} \Bigg( \frac{\log(nn_3)}{q} \|\T{D}_{t-1}\|_\infty \\
&+ \sqrt{\frac{\log(nn_3)}{q}}\|\T{D}_{t-1}\|_{\infty,2^*}  \Bigg) \\
\le &\frac{c}{\sqrt{c_0}}\sum_{t=1}^{t_0}\left( \frac{n}{\mu_0r}\|\T{D}_{t-1}\|_\infty + \sqrt{\frac{n}{\mu_0 r}} \|\T{D}_{t-1}\|_{\infty,2^*}  \right)
\end{aligned}
\end{equation}
where we could bound term $\|\T{D}_{t-1}\|_\infty$ using \textbf{Lemma} \ref{lemma:4} as follows,
\begin{equation}
\label{eq:pro_tmp_2}
\begin{aligned}
&\|\T{D}_{t-1}\|_\infty\\
= &\|(\T{P}_T - \T{P}_T \T{R}_{\Omega_{t-1}}\T{P}_T )...(\T{P}_T - \T{P}_T \T{R}_{\Omega_1}\T{P}_T )\| \\
\le & \left( \frac{1}{2} \right)^{t-1} \|\TU*\TV\Tra\|_\infty
\end{aligned}
\end{equation}
and $\|\T{D}_{t-1}\|_{\infty,2^*}$ is bounded using \textbf{Lemma}~\ref{lemma:3} and (\ref{eq:pro_tmp_1})(\ref{eq:pro_tmp_2}),
\begin{equation}
\nonumber
\begin{aligned}
&\|\T{D}_{t-1}\|_{\infty,2^*}\\
= &\left\| (\T{P}_T - \T{P}_T \T{R}_{\Omega_{k-1}} \T{P}_T )(\T{D}_{t-2}) \right\|_{\infty,2^*} \\
\le &  \frac{1}{2}\left\|\T{D}_{t-2} \right\|_{\infty,2^*} + \frac{1}{2}\sqrt{\frac{n}{\mu_0 r}} \left\| \T{D}_{k-2} \right\|_\infty \\
\le & \frac{1}{2}\left(  \frac{1}{2}\left\|\T{D}_{t-3} \right\|_{\infty,2^*} + \frac{1}{2}\sqrt{\frac{n}{\mu_0 r}} \left\| \T{D}_{k-3} \right\|_\infty \right) \\
&+ \frac{1}{2}\sqrt{\frac{n}{\mu_0 r}} \left\| \T{D}_{k-2} \right\|_\infty \\
\le & ... \\
\le & t\left( \frac{1}{2} \right)^{t-1} \sqrt{\frac{n}{\mu_0 r}} \left\| \TU*\TV\Tra \right\|_\infty \\
&+ \left( \frac{1}{2} \right)^{t-1} \left\| \TU*\TV\Tra \right\|_{\infty,2^*}
\end{aligned}
\end{equation}

So we get
\begin{equation}
\nonumber
\begin{aligned}
&\left\| \T{P}_{T^\perp} (\T{Y}) \right\| \\
\le &\frac{c}{\sqrt{c_0}} \frac{n}{\mu_0 r}\|\TU*\TV\Tra\|_\infty \sum_{t=1}^{t_0}(t+1)\left( \frac{1}{2}\right)^{t-1} \\
&+ \frac{c}{\sqrt{c_0}}\sqrt{\frac{n}{\mu_0 r}} \|\TU*\TV\Tra\|_{\infty,2^*} \sum_{t=1}^{t_0} \left( \frac{1}{2} \right)^{t-1} \\
\le & \frac{6c}{\sqrt{c_0}} \frac{n}{\mu_0 r}\|\TU*\TV\Tra\|_\infty + \frac{2c}{\sqrt{c_0}}\sqrt{\frac{n}{\mu_0r}} \left\| \TU*\TV\Tra \right\|_{\infty,2^*}
\end{aligned}
\end{equation}

First let's bound $\|\TU*\TV\Tra\|_\infty$. We have

\begin{equation}
	\nonumber
	\begin{aligned}
		\|\TU*\TV\Tra\|_\infty = &\max_{i,j,k}\left( \TU(i,:,:) * \TV\Tra(:,j,:) \right)_k \\
		= & \max_{i,j} \| \TU(i,:,:) * \TV\Tra(:,j,:) \|_\infty 
	\end{aligned}
\end{equation}

Note the fact that for two tensor tubes $\tube{x},\tube{y} \in \mathbb{R}^{1\times 1 \times n_3}$, use the Cauchy-Schwartz inequality we get
%
\begin{equation}
\nonumber
\|\tube{x}* \tube{y}\|_\infty \le \|\tube{x}\|_{2^*} \|\tube{y}\|_{2^*}
\end{equation}

Then let $\tube{u}_t = \TU(i,t,:), \tube{v}_t\Tra = \TV\Tra(t,j,:)$, we can further write $\|\TU*\TV\Tra\|_\infty$ as follows
\begin{equation}
\nonumber
\begin{aligned}
\|\TU*\TV\Tra\|_\infty = & \max_{i,j} \left\| \sum_{t=1}^{r} \tube{u}_t * \tube{v}_t\Tra   \right\|_\infty \\
\le & \max_{i,j} \sum_{t=1}^{r} \| \tube{u}_t * \tube{v}_t\Tra  \|_\infty \\
\le & \max_{i,j} \sum_{t=1}^{r} \|\tube{u}_t \|_{2^*} \|\tube{v}_t\Tra\|_{2^*} \\
\le & \max_{i,j} \sum_{t=1}^{r} \frac{1}{2} \left( \|\tube{u}_t \|_{2^*}^2 + \|\tube{v}_t\Tra\|_{2^*}^2 \right) \\
= & \max_{i,j} \left\{ \frac{1}{2}\|\tcol{e}_i\Tra * \TU\|_{2^*}^2 + \frac{1}{2} \|\TV\Tra * \tcol{e}_j\|_{2^*}^2 \right\}\\
\le &\frac{\mu_0 r}{n}
\end{aligned}
\end{equation} 
by the standard incoherent condition. We also have
\begin{equation}
\nonumber
\begin{aligned}
&\|\TU*\TV\Tra\|_{\infty,2^*}\\
= &\max_{i,j} \left\{ \|\TU*\TV\Tra * \tcol{e}_i \|_{2^*}, \|\tcol{e}_j\Tra * \TU * \TV\Tra\|_{2^*} \right\} \\
\le & \sqrt{\frac{\mu_0r}{n}}
\end{aligned}
\end{equation}
and thus
\begin{equation}
\nonumber
\|\T{P}_{T^\perp} (\T{Y})\| \le \frac{8c}{\sqrt{c_0}} \le \frac{1}{2}
\end{equation}
given $c_0$ large enough. 

\end{proof}


\begin{IEEEbiography}{Zemin Zhang}
received the BSc degree in electronics information engineering from Tsinghua University, China, in 2007. He is currently working toward the PhD degree in Electrical Engineering and Computer Science at Tufts University under the supervision of Prof. Shuchin Aeron. His research interests include computer vision, machine learning, and tensor analysis and applications.

\end{IEEEbiography}

\begin{IEEEbiography}{Shuchin Aeron}
	Shuchin Aeron is currently an assistant professor in the department of ECE at Tufts University. He obtained his MS and PhD degrees from Boston University in 2004 and 2009 respectively. His research interests lie in information theory and statistical signal processing, and their applications to computer vision and geophysical signal processing.  
\end{IEEEbiography}

\newpage
\noindent \textbf{Supplementary Proofs}\\
\\
\\
\noindent \textbf{Proof of Lemma C.1}
\begin{proof}
	Let
	\begin{equation}
		\nonumber
		\begin{aligned}
			&\T{R}_\Omega(\T{Z}) - \T{Z} \\
			= &\sum_{i,j,k}\T{C}_{(ijk)} \\
			=& \sum_{i,j,k}\left( \frac{1}{p}\delta_{ijk} - 1  \right) \T{Z}_{ijk} \tcol{e}_i *\tube{e}_k *\tcol{e}_j\Tra
		\end{aligned}
	\end{equation}
	\noindent where $\T{C}_{(ijk)}$ are independent tensors. Then we have
	\begin{equation}
		\nonumber
		\xbar{\T{C}_{(ijk)}} = \sum_{i,j,k}\left( \frac{1}{p}\delta_{ijk} - 1  \right) \T{Z}_{ijk} \xbar{\tcol{e}_i}\xbar{\tube{e}_k}\xbar{\tcol{e}_j}\Tra
	\end{equation}
	Notice that $\mathbb{E}\left[ \xbar{\T{C}_{(ijk)}}  \right]=0$ and 
	$\|\xbar{\T{C}_{(ijk)}}\| \le \frac{1}{p} \|\T{Z}\|_{\infty}$. Moreover,
	\begin{equation}
		\nonumber
		\begin{aligned}
			&\left\| \mathbb{E}\left[  \sum_{i,j,k}\xbar{\T{C}_{(ijk)}}\Tra\xbar{\T{C}_{(ijk)}} \right] \right\| \\
			= &\left\| \mathbb{E}\left[  \sum_{i,j,k}\T{C}_{(ijk)}\Tra\T{C}_{(ijk)} \right] \right\| \\
			= & \left\| \sum_{i,j,k} \T{Z}_{ijk}^2 \tcol{e}_j * \tcol{e}_j\Tra  \mathbb{E}\left( \frac{1}{p}\delta_{ijk} - 1  \right)^2 \right\| \\
			= &\left\| \frac{1-p}{p}\sum_{i,j,k} \T{Z}_{ijk}^2 \tcol{e}_j * \tcol{e}_j\Tra \right\| 
		\end{aligned}
	\end{equation}
	\noindent since $\tcol{e}_j * \tcol{e}_j \Tra$ will return a zero tensor except for $(j,j,1)$th entry equaling $1$, we have
	\begin{equation}
		\nonumber
		\begin{aligned}
			&\left\| \mathbb{E}\left[  \sum_{i,j,k}\xbar{\T{C}_{(ijk)}}\Tra\xbar{\T{C}_{(ijk)}} \right] \right\| \\
			= & \frac{1-p}{p} \max_{j} \left| \sum_{i,k} \T{Z}_{ijk}\right| \\
			\le & \frac{1}{p} \|\T{Z}\|_{\infty,2^*}^2
		\end{aligned}
	\end{equation}
	\noindent And $\left\| \mathbb{E}\left[  \sum_{i,j,k}\xbar{\T{C}_{(ijk)}}\xbar{\T{C}_{(ijk)}}\Tra \right] \right\| $ is bounded similarly. Then use the extension of \textbf{Theorem A.1}, for any $c'>0$ we have
	\begin{equation}
		\nonumber
		\begin{aligned}
			&\|\T{R}_\Omega(\T{Z}) - \T{Z}\| = \|\xbar{\T{R}_\Omega(\T{Z})} - \xbar{\T{Z}}\| \\
			= &\left\| \sum_{i,j,k}\xbar{\T{C}_{(ijk)}} \right\| \\
			\le & \sqrt{\frac{4c'}{p}\|\T{Z}\|^2_{\infty,2^*} \log (2nn_3)} + \frac{c'}{p}\|\T{Z}\|_\infty \log (2nn_3) \\
			\le &c\left( \frac{\log (nn_3)}{p} \|\T{Z}\|_\infty + \sqrt{\frac{\log (nn_3)}{p}}\|\T{Z}\|_{\infty,2^*} \right)
		\end{aligned}
	\end{equation}
	holds with probability at lease $1-(2nn_3)^{-(c-1)}$ for any $c \ge \max \{ c' , 2\sqrt{c'} \}$.
\end{proof}


\noindent \textbf{Proof of Lemma C.2}

\begin{proof}
	Consider any $b$th tensor column of $\T{P}_T \T{R}_\Omega (\T{Z}) - \T{P}_T(\T{Z})$:
	\begin{equation}
	\nonumber
	\begin{aligned}
	&\left( \T{P}_T \T{R}_\Omega (\T{Z}) - \T{P}_T(\T{Z}) \right) * \tcol{e}_b \\
	=& \sum_{i,j,k}(\frac{1}{p}\delta_{ijk} - 1)\T{Z}_{ijk} \T{P}_T (\tcol{e}_i* \tube{e}_k * \tcol{e}_j\Tra)*\tcol{e}_b \\
	:= &\sum_{i,j,k} \tcol{a}_{ijk}
	\end{aligned}
	\end{equation}
	where $\tcol{a}_{ijk} \in \mathbb{R}^{n \times 1 \times n_3}$ are zero-mean independent tensor columns. Let $\tvec{a}_{ijk} \in \mathbb{R}^{nn_3 \times 1}$ be the vectorized column vector of $\tcol{a}_{ijk}$. Then the $\ell_2$ norm of the vector $\tvec{a}_{ijk}$ is bounded by the following
	\begin{equation}
	\nonumber
	\begin{aligned}
	&\|\tvec{a}_{ijk}\| \\
	= &\|\tcol{a}_{ijk}\|_{2^*} \\
	\le &\frac{1-p}{p}\T{Z}_{ijk} \left\|\T{P}_T (\tcol{e}_i* \tube{e}_k * \tcol{e}_j\Tra)*\tcol{e}_b\right\|_{2^*} \\
	\le &\frac{1}{p} \sqrt{\frac{2\mu_0 r}{n}}\|\T{Z}\|_\infty \\
	\le &\frac{1}{c_0\log(nn_3)} \sqrt{\frac{2n}{\mu_0 r}}\|\T{Z}\|_\infty
	\end{aligned}
	\end{equation}
	for some constant $c_0 > 0$ given $p$ satisfying (\ref{eq:p}). We also have
	\begin{equation}
	\nonumber
	\begin{aligned}
	&\left| \mathbb{E}\left[ \sum_{i,j,k} \tvec{a}_{ijk}\Tra \tvec{a}_{ijk}  \right]   \right| \\
	= &\mathbb{E}\left[ \sum_{i,j,k} \|\tcol{a}_{ijk}\|_{2^*}^2  \right]  \\
	=&\frac{1-p}{p} \sum_{i,j,k} \T{Z}_{ijk}^2 \left\|\T{P}_T (\tcol{e}_i* \tube{e}_k * \tcol{e}_j\Tra)*\tcol{e}_b \right\|_{2^*}^2
	\end{aligned}
	\end{equation}
	Use the definition of $\T{P}_T$ and the incoherent condition, we can write
	\begin{equation}
	\nonumber
	\begin{aligned}
	&\Big\|\T{P}_T (\tcol{e}_i* \tube{e}_k * \tcol{e}_j\Tra)*\tcol{e}_b\Big\|_{2^*} \\
	= &\Big\| ( \TU * \TU\Tra * \tcol{e}_i *\tube{e}_k ) * \tube{e}_j\Tra *\tcol{e}_b  \\
	& + (\T{I}-\TU*\TU\Tra)*\tcol{e}_i *\tube{e}_k * \tube{e}_j\Tra * \TV * \TV\Tra * \tcol{e}_b    \Big\|_{2^*}\\
	\le & \sqrt{\frac{\mu_0 r}{n}} \Big\| \tcol{e}_j\Tra * \tcol{e}_b \Big\|_{2^*} \\
	& + \Big\| (\T{I} - \TU*\TU\Tra )*\tcol{e}_i*\tcol{e}_k  \Big\| \Big\| \tcol{e}_j\Tra * \TV * \TV\Tra * \tcol{e}_b  \Big\|_{2^*} \\
	\le & \sqrt{\frac{\mu_0 r}{n}} \Big\| \tcol{e}_j\Tra * \tcol{e}_b \Big\|_{2^*} + \Big\| \tcol{e}_j\Tra * \TV * \TV\Tra * \tcol{e}_b  \Big\|_{2^*}
	\end{aligned}
	\end{equation}
	where $\T{I}$ is the identity tensor. Thus,
	\begin{align}
	\nonumber
	&\left| \mathbb{E}\left[ \sum_{i,j,k} \tvec{a}_{ijk}\Tra\tvec{a}_{ijk} \right]    \right| \\
	\nonumber
	\le & \frac{2}{p} \sum_{ijk} \T{Z}_{ijk}^2 \frac{\mu_0 r}{n} \left\| \tcol{e}_j\Tra * \tcol{e}_b \right\|_{2^*}^2 \\
	\nonumber
	&+ \frac{2}{p}\sum_{ijk}\T{Z}_{ijk}^2 \left\| \tcol{e}_j\Tra * \TV * \TV\Tra * \tcol{e}_b \right\|_{2^*}^2  \\
	\nonumber
	= & \frac{2\mu_0 r}{pn}\sum_{i,k} \T{Z}_{ibk}^2 \\
	\label{eq:a21}
	&+ \frac{2}{p} \sum_{j} \left\| \tcol{e}_j\Tra * \TV * \TV\Tra * \tcol{e}_b \right\|_{2^*}^2 \sum_{i,k}\T{Z}_{ijk}^2 \\
	\nonumber
	\le & \frac{2\mu_0 r}{pn} \left\| \T{Z}\right\|^2_{\infty,2^*} + \frac{2}{p} \left\| \TV * \TV \Tra * \tcol{e}_b \right\|^2_{2^*} \left\| \T{Z}\right\|^2_{\infty,2^*} \\
	\nonumber
	\le & \frac{4\mu_0 r}{pn} \|\T{Z}\|^2_{\infty,2^*} \\
	\nonumber
	\le & \frac{4}{c_0 \log(nn_3)}
	\end{align}
	where (\ref{eq:a21}) is because $\tcol{e}_j\Tra * \tcol{e}_b = 0$ if $j \neq b$. In the same fashion $\left| \mathbb{E}\left[ \sum_{i,j,k} \tvec{a}_{ijk}\tvec{a}_{ijk}\Tra \right] \right|$ is bounded by the exact same quantity. Since the spectral norm of the vector $\tvec{a}_{ijk}$ is equal to its $\ell_2$ norm, then use the extension of \textbf{Theorem} \ref{thm:NBI} we have for any $c_1 > 0$, we have
	\begin{equation}
	\nonumber
	\begin{aligned}
	& \left\| \left( \T{P}_T \T{R}_\Omega (\T{Z}) - \T{P}_T(\T{Z}) \right) * \tcol{e}_b \right\|_{2^*} \\
	= & \left\| \sum_{i,j,k} \tcol{a}_{ijk} \right\|_{2^*} \\
	= &\left\| \sum_{i,j,k} \tvec{a}_{ijk} \right\| \\
	\le &\sqrt{4c_1 \sigma^2 \log(nn_3)} + c_1 M \log(nn_3) \\
	\le &\frac{1}{2}\left\| \T{Z} \right\|_{\infty,2^*} + \frac{1}{2}\sqrt{\frac{n}{\mu_0 r}} \left\| \T{Z} \right\|_\infty
	\end{aligned}
	\end{equation}
	holds with probability at least $1-(nn_3)^{-(c_2-1)}$ for $c_2$ large enough.
	
	We can also do the same to the tensor rows $\tcol{e}_a\Tra* \left( \T{P}_T \T{R}_\Omega (\T{Z}) - \T{P}_T(\T{Z}) \right)$ and get the same bound. Then using a union bound over all the tensor columns and tensor rows, the result holds with probability at least $1-2n^2 n_3^{-(c_2-1)}$. With $c_2$ large enough the probability goes to zero which finishes the proof.
\end{proof}


\noindent \textbf{Proof of Lemma C.3}

\begin{proof}
	Observe that
	\begin{equation}
	\nonumber
	\T{P}_T \T{R}_\Omega \T{P}_T ( \T{Z} ) = \sum_{i,j,k} \frac{1}{p}\delta_{ijk}\T{Z}_{ijk}\T{P}_T(\tcol{e}_i * \tube{e}_k * \tcol{e}_j\Tra)
	\end{equation}
	so we have that any $(a,b,c)$th entry of $\T{P}_T \T{R}_\Omega \T{P}_T ( \T{Z} ) - \T{P}_T ( \T{Z} )$ is given by
	\begin{equation}
	\nonumber
	\begin{aligned}
	&\langle \T{P}_T \T{R}_\Omega \T{P}_T ( \T{Z} ) - \T{P}_T ( \T{Z} ) , \tcol{e}_a * \tube{e}_c * \tcol{e}_b\Tra \rangle \\
	= &\sum_{i,j,k} \left( \frac{\delta_{ijk}}{p}-1\right) \T{Z}_{ijk}\langle \T{P}_T(\tcol{e}_i * \tube{e}_k * \tcol{e}_j\Tra),\tcol{e}_a * \tube{e}_c * \tcol{e}_b\Tra\rangle \\
	:= & \sum_{i,j,k} \T{H}_{ijk}
	\end{aligned}
	\end{equation}
	
	It is easy to observe that 
	\begin{equation}
	\nonumber
	\begin{aligned}
	&\left| \T{H}_{ijk} \right| \\
	\le & \frac{1}{p}\|\T{Z}\|_\infty \|\T{P}_T(\tcol{e}_i * \tube{e}_k * \tcol{e}_j\Tra)\|_F \|\T{P}_T(\tcol{e}_a * \tube{e}_c * \tcol{e}_b\Tra)\|_F \\
	\le & \frac{2\mu_0 r}{np}\|\T{Z}\|_\infty
	\end{aligned}
	\end{equation}
	
	We also have
	\begin{equation}
	\nonumber
	\begin{aligned}
	&\left| \mathbb{E} \left[ \sum_{i,j,k} \T{H}_{ijk}^2 \right] \right| \\
	= & \frac{1-p}{p} \|\T{Z}\|_\infty^2\sum_{i,j,k}  \left| \langle \T{P}_T(\tcol{e}_i * \tube{e}_k * \tcol{e}_j\Tra),\tcol{e}_a * \tube{e}_c * \tcol{e}_b\Tra\rangle \right|^2 \\
	= & \frac{1-p}{p} \|\T{Z}\|_\infty^2 \left\| \T{P}_T (\tcol{e}_a * \tube{e}_c * \tcol{e}_b\Tra )  \right\|_F \\
	\le & \frac{2\mu_0r}{np} \|\T{Z}\|_\infty^2
	\end{aligned}
	\end{equation}
	
	Then use \textbf{Theorem A.1}, we have 
	\begin{equation}
	\nonumber
	\begin{aligned}
	&\mathbb{P}\left[ \left(\T{P}_T \T{R}_\Omega \T{P}_T ( \T{Z} ) - \T{P}_T ( \T{Z} )\right)_{abc} \ge \frac{1}{2}\|\T{Z}\|_\infty \right] \\
	\le &2 \exp \left( \frac{-\|\T{Z}\|_\infty^2/4}{ \frac{2\mu_0r}{np}\|\T{Z}\|_\infty^2 + \frac{\mu_0r}{3np}\|\T{Z}\|_\infty^2   } \right)\\
	\le &2(nn_3)^{-c_3}
	\end{aligned}
	\end{equation}
	for some $c_3 = 3c_0/28$ large enough given $p$ satisfying (24). Then using the union bound on every $(a,b,c)$th entry we have $\| (\T{P}_T\T{R}_{\Omega} \T{P}_T - \T{P}_T ) (\T{Z}) \|_\infty \le \frac{1}{2} \|\T{Z}\|_\infty$ holds with probability at least $1-2n^{-(c_3-2)}n_3^{-(c_3-1)}$.
	
\end{proof}


%
%
%



%


\end{document}